\documentclass[acmtog]{acmart}
\acmSubmissionID{162}

\usepackage{booktabs} 
\usepackage{soul} 
\usepackage{enumitem} 

\usepackage{graphicx}
\usepackage{subcaption}
\usepackage{mwe}

\newcommand{\modelName}{\textit{BrepGen}}
\newcommand{\datasetName}{\textit{Furniture B-rep Dataset}}

\usepackage[ruled]{algorithm2e} 

\SetAlFnt{\small}
\SetAlCapFnt{\small}
\SetAlCapNameFnt{\small}
\SetAlCapHSkip{0pt}

\acmJournal{TOG}

\begin{document}
\title{\modelName: A B-rep Generative Diffusion Model with Structured Latent Geometry}

\author{Xiang Xu}
\orcid{0000-0002-3437-1470}
\affiliation{%
 \institution{Simon Fraser University}
 \country{Canada}}
\affiliation{%
 \institution{Autodesk Research}
 \country{Canada}}
\email{xuxiangx@sfu.ca}

\author{Joseph G. Lambourne}
\orcid{0000-0002-9892-1945}
\affiliation{%
 \institution{Autodesk Research}
 \country{UK}}
\email{joseph.lambourne@autodesk.com}

\author{Pradeep Kumar Jayaraman}
\orcid{0000-0001-6314-6136}
\affiliation{%
 \institution{Autodesk Research}
 \country{Canada}}
\email{pradeep.kumar.jayaraman@autodesk.com}

\author{Zhengqing Wang}
\orcid{0009-0004-8067-1927}
\affiliation{%
 \institution{Simon Fraser University}
 \country{Canada}}
\email{zwa170@sfu.ca}

\author{Karl D.D. Willis}
\orcid{0000-0002-6990-2294}
\affiliation{%
 \institution{Autodesk Research}
 \country{USA}}
\email{karl.willis@autodesk.com}

\author{Yasutaka Furukawa}
\orcid{0009-0006-9775-4512}
\affiliation{%
 \institution{Simon Fraser University}
 \country{Canada}}
\email{furukawa@sfu.ca}

\begin{abstract}
This paper presents \modelName, a diffusion-based generative approach that directly outputs a Boundary representation (B-rep) Computer-Aided Design (CAD) model. 
\modelName\ represents a B-rep model as a novel structured latent geometry in a hierarchical tree. With the root node representing a whole CAD solid, each element of a B-rep model (i.e., a face, an edge, or a vertex) progressively turns into a child-node from top to bottom. B-rep geometry information goes into the nodes as the global bounding box of each primitive along with a latent code describing the local geometric shape. The B-rep topology information is implicitly represented by node duplication. When two faces share an edge, the edge curve will appear twice in the tree, and a T-junction vertex with three incident edges appears six times in the tree with identical node features. Starting from the root and progressing to the leaf, \modelName\ employs Transformer-based diffusion models to sequentially denoise node features while duplicated nodes are detected and merged, recovering the B-Rep topology information. Extensive experiments show that \modelName\ advances the task of CAD B-rep generation, surpassing existing methods on various benchmarks. Results on our newly collected furniture dataset further showcase its exceptional capability in generating complicated geometry. While previous methods were limited to generating simple prismatic shapes, \modelName\ incorporates free-form and doubly-curved surfaces for the first time. Additional applications of \modelName\ include CAD autocomplete and design interpolation. The code, pretrained models, and dataset are available at \url{https://github.com/samxuxiang/BrepGen}.

\end{abstract}

%
%


%
%


\begin{teaserfigure}
\includegraphics[width=0.99\textwidth]{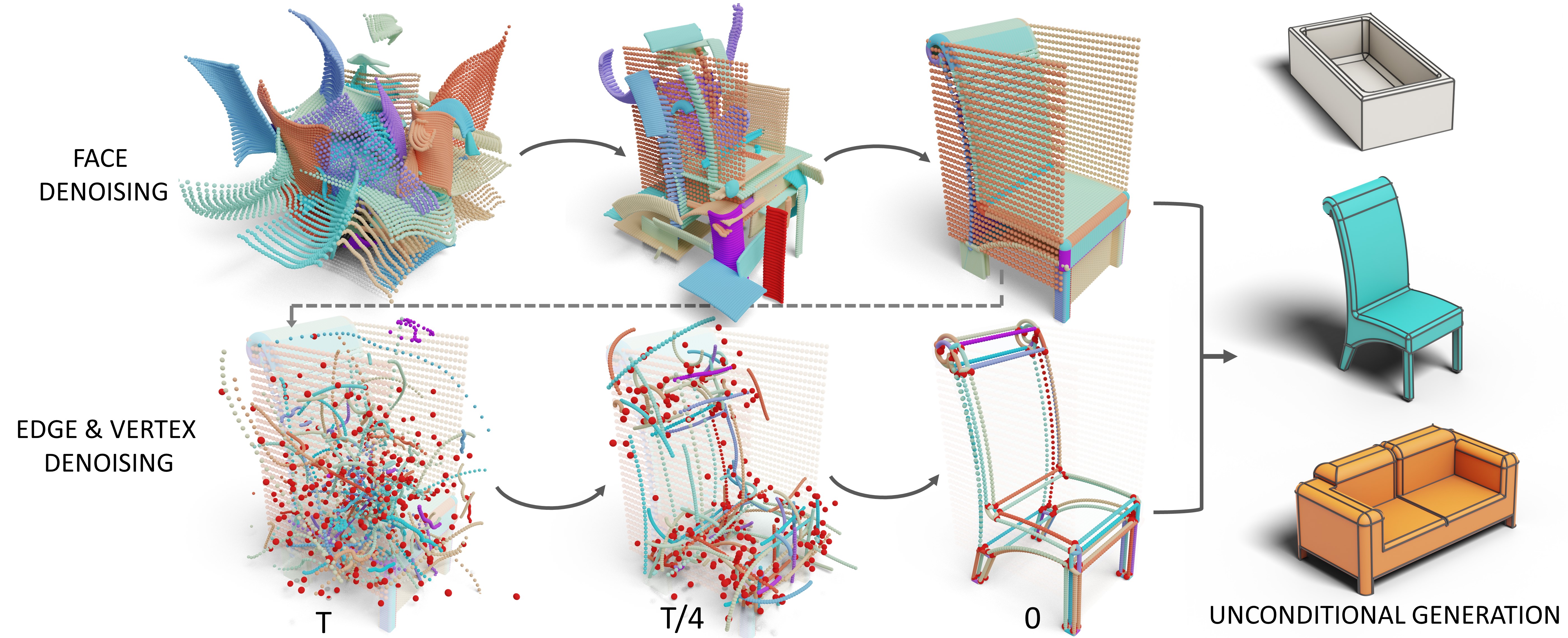}
  \caption{\modelName\ is a diffusion-based approach to generate 3D CAD models in B-rep format. Geometry and topology relations are progressively denoised for faces, edges, and vertices to form a water-tight solid at the end. Extensive evaluations demonstrate that \modelName\ generates a wide range of complex B-reps.  
  }
  \label{fig:teaser}
\end{teaserfigure}

\setcopyright{acmlicensed}
\acmJournal{TOG}
\acmYear{2024} \acmVolume{43} \acmNumber{4} \acmArticle{119} \acmMonth{7}\acmDOI{10.1145/3658129}

\maketitle

\section{Introduction}
\label{sec:intro}

Nearly every man-made object begins its life as a Computer-Aided Design (CAD) model. Within the realm of CAD, the Boundary representation (B-rep) is the predominant format for describing shapes, and is widely used in freeform surface modeling to express complicated geometry. The B-rep representation \cite{weiler1986} generalizes half-edge meshes \cite{MULLER1978}, allowing parametric curves and surfaces to replace the planar facets and linear edges used in mesh modeling. A B-rep consists of sets of interconnected faces, edges, and vertices. A face is the visible region of a parametric surface, bounded by closed loops formed by its adjacent edges, while an edge is the visible region of a parametric curve, trimmed by vertices that define its start and end points. The adjacency of neighbouring edges and vertices is recorded, allowing the structure to provide a complete description of the final solid shape. A system capable of directly generating B-reps would revolutionize the CAD design workflow, reducing the extensive manual labor required from skilled designers and the reliance on professional CAD software.

However, directly generating the B-rep poses significant challenges. In contrast to triangle meshes, B-reps contain a variety of parametric surface and curve types such as arcs, toruses, and Non-Uniform Rational B-Splines (NURBS). Each geometry has a different definition and its own set of parameters, making it difficult to generate. Furthermore, the topological relations between all geometry must be correct to form a water-tight solid. Recent CAD generative models avoid direct B-rep generation and instead focus on sketch and extrude modeling operations \cite{wu2021deepcad, xu2022skexgen, xy2023Hierarchical, ren2022extrude, Li2023SECADNet, Zhou2023CADParser} that represent only a limited range of 3D shapes. Direct B-rep generation methods, like SolidGen \cite{jayaraman2022solidgen}, are limited to highly simplified prismatic shapes and non-freeform surfaces. 

To this end, we introduce \modelName, a generative approach that directly outputs diverse B-rep CAD models using Denoising Diffusion Probabilistic Models (DDPM)~\cite{ho2020denoising}. Key to our method is the use of a structured latent geometry representation that transforms any B-rep into a tree data structure. Concretely, \modelName\ encodes the geometry as node features in a tree, where the root node identifies the CAD solid, and child-nodes, at the following three levels, define the global position and latent local geometry for every face, edge, and vertex. For the local geometry, we follow UV-Net \cite{jayaraman2021uvnet} and use a grid of points uniformly-sampled along the UV domain as a substitute for its parametric function. Mating and association topologies are encoded through the duplication of tree nodes in two ways. First, every shared edge (resp. vertex) in a B-rep model turns into multiple nodes in a tree with the same node features, establishing connections between adjacent geometries and their corresponding mating duplicates. Second, since the number of faces constituting a solid and the number of edges bounding a face vary, an additional duplication is performed to 
pad child nodes at each parent to a predefined length, leading to uniform width trees.

This node duplication is the key in our structured latent geometry representation, encoding both the B-rep geometry and topology information in a unified tree format, where continuous geometry regression implicitly recovers discrete topology. Concretely, a Transformer-based diffusion model generates node features top to bottom (i.e., face to edge to vertex), while nodes with similar features are detected and merged 1) across different parents to recover mating and adjacency relations, and 2) within each parent to restore unique geometric elements associated with it. 
By merging the duplicated nodes, the topology of each face is reconstructed, first connecting edges into closed loops and then using them to trim the parent face. The resulting trimmed faces are joined together to directly output the CAD model in B-rep format. In summary, we make the following contributions:

\begin{itemize}[leftmargin=.25in]
  \item A structured latent geometry representation, whose hierarchical tree, with node duplication, encodes the B-rep geometry and topology information in a unified format. 
  \item A latent diffusion module, capable of generating free-form surfaces and trimming curves.
  \item A newly collected \datasetName\ with high-quality B-reps of indoor objects across 10 different categories.
  \item Direct B-rep generation with state-of-the-art performance.
\end{itemize}

\begin{figure*}[ht]
    \begin{center}
    \includegraphics[width=0.99\textwidth]{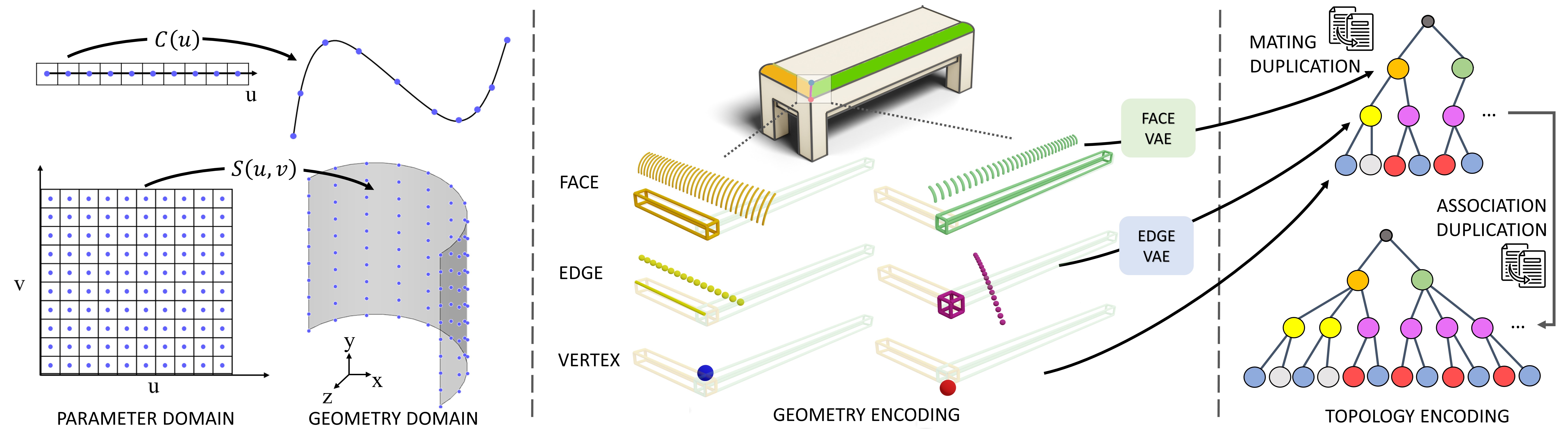}
        \caption{Our structured latent geometry tree representation of a B-rep CAD model. Shape feature is sampled from the parametric surface or curve using grid in the parameter domain (left). Position and local geometry are extracted for the face, edge, and vertex, and then encoded as node features in the tree (mid). Topology is encoded by mating and association duplication. Node feature is represented by color values, where duplicated nodes are of the same color (right).}
        \label{fig:data}
    \end{center}
\end{figure*}

\section{Related Work}
\label{sec:related_work}

This section reviews three CAD model representations used in the literature: constructive solid geometry, sketch and extrude, and B-rep, followed by a summary of B-rep datasets and the use of diffusion models for CAD model generation.

\subsection{Constructive Solid Geometry (CSG)} CSG is a common method to create 3D shapes by combining primitives (e.g. cuboids, spheres) with Boolean operations (e.g. union, subtraction) to form a CSG tree. CSG has been used extensively in `shape programs'~\cite{ritchie2023neurosymbolic} both with neural guidance~\citep{sharma2018csgnet,ellis2019write,tian2019learning} and without~\citep{du2018inversecsg,nandi2017programming,nandi2018functional}, thanks to its simplicity. Recent CSG-based approaches have focused on recovering 3D shapes as primitive assemblies without supervision from the ground truth CSG tree~\cite{kania2020ucsg, yu2022capri, ren2021csg, yu2023dualcsg}. Even though most CSG primitives can be represented as B-reps, converting non-trivial CSG shapes into B-reps induces unwanted complexity. For example, thin sliver faces are created when CSG primitives are almost aligned but not coincident.
A major advantage of our work is that we learn from the B-rep topology found in industrial quality designs, resulting in well structured B-rep output that can be edited using standard direct modeling CAD tools.

\subsection{Sketch and Extrude}
Sketch and extrude is a popular format, representing a CAD model as a sequence of modeling operations stored in parametric CAD files. Reconstruction~\citep{Li2023SECADNet, xu2021zone} and generation~\citep{wu2021deepcad,xu2022skexgen,ren2022extrude, Zhou2023CADParser, xy2023Hierarchical} of the format have been studied in recent years. While significant progress has been made on improving the fidelity and controllability, sequential CAD generative models currently are limited to producing sketches comprised of line, arc and circle primitives, and only the extrude modeling operation. Our method focuses on direct synthesis of the B-rep data with emphasis on supporting complex curves and surfaces that cannot be achieved by previous methods.

\subsection{Boundary Representation}
B-rep 3D models are represented as a graph~\citep{ansaldi1985geometric}, consisting of geometric primitives i.e., parametric curves and surfaces, and topological primitives, i.e., vertices, edges, and faces that are used to trim and sew the surface patches into a solid model. B-rep classification and segmentation tasks have been tackled with graph neural networks~\citep{cao2020gnn, jayaraman2021uvnet, willis2022joinable}, custom convolutions~\citep{lambourne2021brepnet}, and hierarchical graph structures~\citep{jones2021automate, bian2023hgcad, jones2023self}.
For generation, previous approaches handled predefined template shapes~\citep{smirnov2021patches} such as parametric curves~\citep{wang2020pie} and surfaces~\citep{sharma2020parsenet,li2019supervised}. A challenge remains for cases, where a watertight solid is formed by trimming and connecting surfaces.

PolyGen~\citep{nash2020polygen} uses Transformers~\citep{vaswani2017transformer} and pointer networks~\citep{vinyals2015pointer} to generate n-gon meshes, which is a special case of B-rep models with planar faces and linear edges. Other mesh generation methods also include BSP-Net~\citep{chen2020bspnet} and MeshGPT~\citep{siddiqui2023meshgpt}. Wang \textit{et al.} reconstructs a B-rep model from a 2D wireframe drawing by identifying planar and cylindrical faces~\cite{wang2022neuralface}. Guo \textit{et al.} reconstructs B-rep from point clouds using a neural network to predict the rough B-rep geometry and topology, that were refined using a combinatorial optimization~\citep{guo2022complexgen}. 

Closest to our work is SolidGen~\cite{jayaraman2022solidgen}, that can generate entire B-reps, first with the synthesis of vertices, followed by conditional construction of edge topology using a pointer network, finally followed by another pointer network that conditionally connected the edges to faces. While this approach can produce plausible B-reps, it is restricted to prismatic primitives which limit the complexity of the results, and the amount of data that it can be trained on. Our method is more general and can produce prismatic-looking, as well as freeform doubly-curved geometry.

\subsection{Diffusion Models for CAD Generation}
Diffusion models have been successful at generating geometry when topology is given. HouseDiffusion~\citep{shabani2023housediffusion} generates 2D floor-plan that follows a given room connectivity. CAGE~\citep{liu2023cage} generates 3D articulated object conditioned on the ground-truth part connections. More recent methods like PolyDiff~\citep{alliegro2023polydiff} do not require topology as input conditioning but are also limited to generating triangle soups of polygonal meshes without any topology. Using diffusion model to unconditionally generate well-structured geometry together with the correct topology relations remains a challenge, which is what our method aims to solve.

\subsection{B-rep Datasets}
B-rep 3D datasets have grown in number over the last several years. These include synthetically created datasets with class labels, such as FabWave~\cite{atin2019:fabsearch}, SolidLetters~\cite{jayaraman2021uvnet}, and MFCAD~\cite{cao2020:gnn}, as well as datasets of human designed 3D models without canonical class labels, such as ABC~\cite{koch2019:abc}, Fusion 360 Gallery~\cite{willis2022joinable, lambourne2021brepnet, willis2020fusion}, Automate~\cite{jones2021automate}, and DeepCAD~\cite{wu2021deepcad}. Our \datasetName, to the best of our knowledge, is the first dataset to contain human-designed 3D models in the B-rep format across a standard set of classes (e.g. tables, chairs etc).

\section{Structured Latent Geometry} 
\label{sec:representation}

A B-rep model consists of geometric elements (faces, edges, vertices) with pairwise topological relationships (face-edge, edge-vertex adjacency matrix). The challenge lies in generating these two distinct data representations in a general graph topology. To solve this,  our approach unifies the geometry and topology of a B-rep model as a hierarchical tree with a fixed graph topology, where node features encode the geometry information and duplicated nodes (i.e., nodes with near identical features during generation) implicitly encode the topology information. With the root node representing a whole CAD solid, the tree has three levels: faces, edges, and vertices from top to bottom.
The unified tree representation effectively bridges the two formats and becomes the foundation for training our diffusion models.  This section first explains the node features encoding geometry and then node duplication schemes encoding topology.

\subsection{Geometry Encoding by Node Features}
A face, an edge, or a vertex node encodes its geometry information into a node feature with 1) a global position as bounding box parameters or a point and 2) local shape details as a latent code.

\subsubsection{Face ($F$)} 
Underlying each face is a parametric surface (plane, cylinder, cone, sphere, torus, Bezier, or NURBS) with a function $S(u,v): \mathbb{R}^2 \rightarrow \mathbb{R}^3$, mapping a UV coordinate to a 3D point on the unbounded surface on which the face lies (\autoref{fig:data} left). The shape feature $F_{s}$ is a 2D array of 3D points sampled on the parametric surface, and the position feature $F_{p}$ is the axis-aligned bounding box enclosing the points: $F_{p} = [x_1, y_1, z_1, x_2, y_2, z_2]$ encoding the bottom-left and top-right corners. Following UV-Net~\cite{jayaraman2021uvnet}, let $[u_{\text{min}},  u_{\text{max}}] \times [v_{\text{min}},  v_{\text{max}}] \in \mathbb{R}^2$ be a 2D axis-aligned bounding box along the UV axes. Points are sampled from a $N\times N$ equally spaced grid along the UV axes with step size $\delta u = \frac{u_{\text{max}}-u_{\text{min}}}{N}$, $\delta v = \frac{v_{\text{max}}-v_{\text{min}}}{N}$. 3D coordinates at all grid locations are concatenated to form $F_{s} \in \mathbb{R}^{N \times N \times 3}$ as the shape details. 
Capitalizing on the success of latent diffusion models for image generation~\cite{rombach2022high}, a variational autoencoder (VAE) with a UNet backbone further compresses $F_{s}$ into a latent code $F_z$, after normalizing the 3D coordinates by an affine transformation that maps the bounding box to the canonical cube (i.e., $[-1, 1]^3$). The feature of a face node is then defined as $F = [F_p, F_z]$. Note that this feature does not include the outer trimming boundary or inner holes, which are given by the associated edges. The `Face' row in \autoref{fig:data} (middle) illustrates two faces with different bounding box positions and normalized sampled points, revealing the local shape details.

\subsubsection{Edge ($E$)}
Underlying each edge is a parametric curve (line, circle, elliptic, Bezier, or B-spline) with a function $C(u): \mathbb{R} \rightarrow \mathbb{R}^3$, mapping a $u$-coordinate to a 3D point on the unbounded curve on which the edge lies (\autoref{fig:data} left). The shape feature $E_{s}$ is a 1D array of 3D points sampled along the parametric curve. The position feature $E_{p}$ is again defined as the bounding box parameters enclosing the points. Let $[u_{\text{min}},  u_{\text{max}}] \in \mathbb{R}$ be the minimum and the maximum U-coordinate value of an edge. Points are sampled from an $N$ equally spaced grid along the u-axis with step size $\delta u = \frac{u_{\text{max}}-u_{\text{min}}}{N}$. 3D coordinates at all grid locations are concatenated to form $E_s \in \mathbb{R}^{N \times 3}$ as the shape details. Akin to the face feature construction, a VAE compresses $E_s$ into $E_z$, where $E = [E_p, E_z]$ becomes the feature of an edge node. The `Edge' row in \autoref{fig:data} (middle) visualizes the position and the shape details.

\subsubsection{Vertex ($V$)}
A vertex is a 3D point without additional shape details. The feature of a vertex node is its point coordinate  $V = (x,y,z)$. The `Vertex' row in \autoref{fig:data} (middle) shows the location of two vertices.

\subsection{Topology Encoding by Node Duplication} 

Node duplication in \modelName\ serves two purposes: 1) encode the topological relationships between faces, edges, and vertices, 2) pad the number of faces in a B-rep and edges surrounding a face to a fixed maximum length. Two node duplication schemes achieve these objectives: \textit{mating} and \textit{association}.

\subsubsection{Mating Duplication}
Mating duplication encodes face-edge-face and edge-vertex-edge adjacency relationships geometrically.  Shared edges are duplicated so that each parent face gets a copy of the edge geometry as a child node.   Similarly shared vertices are duplicated as child nodes of their parent edges.  This procedure turns a B-rep graph structure into a tree. The top-right tree in \autoref{fig:data} shows the resulting tree with node features represented by color values, shared blue vertices, and purple duplicated edges. Mating relations are later recovered by merging edge or vertex nodes with similar geometry across different parents.

\subsubsection{Association Duplication}
The number of edges bounding a face and the number of faces forming a solid vary, however the numbers of faces and edges required to build a solid will not be known at inference time. Our idea is to pick a predetermined maximum branching factor for each tree level and randomly duplicate nodes until reaching the maximum number of children. We found this `duplication padding' strategy results in fewer missing faces and edges at inference time than the zero-padding strategy. Randomly selecting the duplicated nodes per training iteration also helps prevent over-fitting. The bottom-right tree in \autoref{fig:data} illustrates this process with the tree randomly padded to two faces and three edges per face. The yellow edge is duplicated once and the purple edge is duplicated twice. Association relations can be recovered by removing child nodes with the same geometry under each parent. Note that every edge is always connected to two vertices of its start and end, making the association duplication unnecessary for vertices.

\begin{figure*}[ht]
    \begin{center}
    \includegraphics[width=0.99\textwidth]{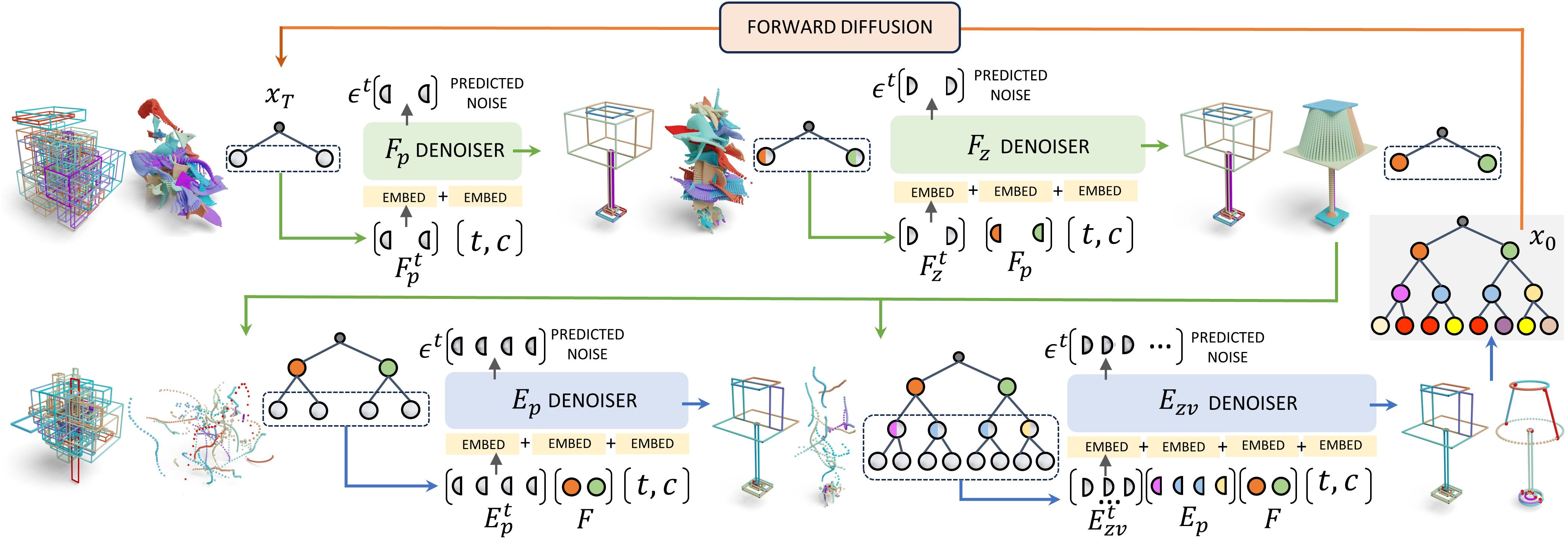}
        \caption{Top row: Generation of the face bounding box position and face latent geometry. Nodes are split in half to represent each. $t$ indicates the time step and $c$ indicates the class label in the Furniture dataset. Bottom row: Generation of the edge bounding box position and edge-vertex joint latent geometry, both conditioned on the parent face. Gray color represents noisy feature values. Points are decoded from the latent geometry for visualization.}
        \label{fig:model}
    \end{center}
\end{figure*}

\section{Method} 
\label{sec:method}
\modelName\ consists of two VAEs that compress the face and edge shape feature into geometry latent vectors, and a latent diffusion module that sequentially denoises the node latent features in a tree from root to leaf, enabling effective training and inference.

\subsection{Shape Geometry VAE}

Following Stable Diffusion \cite{rombach2022high}, we use two VAEs to compress the shape feature $F_s, E_s$ into lower-dimensional latent $F_z, E_z$. Shape feature is first normalized to $[-1,1]$. A VAE with 2D convolution UNet backbone compresses the normalized face feature. A similar VAE with 1D convolution compresses the edge feature. Both VAEs are trained with MSE reconstruction loss and a KL regularization term. Concretely, the face VAE encoder compresses $F_s$ into latent vector $F_z=E(F_s)$. A decoder then reconstructs $\hat{F}_s = D(F_z) = D(E(F_s))$, Reconstruction loss is between $\hat{F}_s$, $F_s$. In practice, we set $N=32$ to densely sample the UV parameter domain, and downsample the input by a factor of $8$ while retaining the feature depth, resulting in $F_z \in \mathbb{R}^{4 \times 4 \times 3}$ and $E_z \in \mathbb{R}^{4 \times 3}$. To further reduce computation, the two endpoint vertices are concatenated with an edge to form joint latent  $E_{zv} \in \mathbb{R}^{4 \times 3 + 6}$.

\subsection{Latent Diffusion Module}
Our latent diffusion module is trained with the compressed latent geometry and global position as node features in the tree. The overall architecture of our generative latent diffusion module is illustrated in \autoref{fig:model}. We follow the DDPM \cite{ho2020denoising} training scheme and use four Transformer-based denoisers to remove the noise added to the nodes sequentially.

\subsubsection{Diffusion Process} 
Given a tree with node features $\mathbf{x}_0$, the forward diffusion process adds Gaussian noise to all the nodes in $\mathbf{T}$ steps. The noisy tree node features $\mathbf{x}_t$ at time step $t$ are sampled as:  
\begin{equation}
\begin{aligned}
q(\mathbf{x}_t \vert \mathbf{x}_0) &= \mathcal{N}(\mathbf{x}_t; \sqrt{\bar{\alpha}_t} \mathbf{x}_0, (1 - \bar{\alpha}_t)\mathbf{I}),\\
\mathbf{x}_t &= \sqrt{\bar{\alpha}_t}\mathbf{x}_0 + \sqrt{1 - \bar{\alpha}_t}\boldsymbol{\epsilon}_t, \\
\end{aligned}
\label{eq:diffusion}
\end{equation}
where noise $\boldsymbol{\epsilon}_t \sim \mathcal{N}(0,\mathbf{I})$ and $\bar{\alpha}_t = \prod_{i=1}^t \alpha_i$, $\alpha_t = 1 - \beta_t
$. The $\beta_t$ is determined by the noise variance scheduler, a linear scheduler by default. After the noise injection, the encoded B-rep geometry and topology are corrupted. For example, duplicated nodes have different features and no longer exhibit the correct topological relations.

\subsubsection{Sequential Denoising} 
Generating all geometry at once is difficult. We use sequential generation to denoise the face, edge, and vertex progressively. Concretely, we factored the distribution of node features $\mathbf{x}$ into a product of conditional distributions:
\begin{equation}
    p(\mathbf{x}) = p(F,E,V) = p(E_{zv}  \vert E_p, F) p(E_p \vert F) p(F_z \vert F_p) p(F_p \vert \emptyset). 
\end{equation}
The ordering reflects a top-down generation process with edge conditioned on face, and latent geometry conditioned on global position. \autoref{fig:model} shows the conditional denoising process with a denoising network for each distribution.

The four denoising networks in our latent diffusion module share a common Transformer backbone. Without loss of generality, we will use edge position denoiser (\autoref{fig:model} bottom-left) as an example. Input to the denoiser are edge tokens with noisy bounding box parameters. Network is conditioned on the previously denoised face tokens from \autoref{fig:model} (top-row), which are embedded as:
\begin{equation}
\label{eq:surf_embed}
    \boldsymbol{F} \leftarrow \text{MLP}(W_p F_p) + \text{MLP}(W_z F_z). 
\end{equation}
$F_p \in \mathbb{R}^{6}$ is the face bounding box coordinates and $F_z \in \mathbb{R}^{48}$ is the flattened latent geometry vector. $W_p \in \mathbb{R}^{d \times 6}$, $W_z \in \mathbb{R}^{d \times 48}$ are the two $d$-dimension embedding matrices. MLP is fully connected layers with SiLU activation. The noisy edge bounding boxes are encoded in a similar manner as $\boldsymbol{E_p} = \text{MLP}(W_p E_p)$.

Rather than using cross-attention, the predefined parent-child relations from the tree are used to directly inject the face condition. Let edge $j$ be the child node of face $i$. Embedding from the $i$th face token is added to the $j$th edge bounding box position token as:
\begin{equation}
\label{eq:edge_embed}
   \boldsymbol{\hat{E}_{p, \hspace{1pt} j}}  \leftarrow  \boldsymbol{E_{p, \hspace{1pt} j}} + \boldsymbol{F_i}.
\end{equation}
During top-down generation, a known number of child nodes are added for each parent at the next stage, leading to simple token addition for the condition. Edge embedding $E_{zv}$ is similarly embedded as $\boldsymbol{E_{zv}} = \text{MLP}(W_{zv} E_{zv})$ with $W_{zv} \in \mathbb{R}^{d \times 18}$. The previously denoised edge position and parent face condition are added to every $j$th edge token as $\boldsymbol{E_{zv, \hspace{1pt} j}} + \boldsymbol{\hat{E}_{p, \hspace{1pt} j}}$ following \autoref{eq:edge_embed}. Final input to the denoiser also adds the time and condition embedding as $f(t,c)$.  Note that we do not use learnable positional encoding.

\subsubsection{Loss Function} 
The four Transformer-based denoisers are trained separately to predict the L2-norm regression loss of the added sampled noise as in DDPM \cite{ho2020denoising}. Loss term is:
\begin{equation}
  L = \mathbb{E}_{t, \mathbf{x}_0, \boldsymbol{\epsilon}_t}\Big[\|\boldsymbol{\epsilon}_t - \boldsymbol{\epsilon}_\theta(\sqrt{\bar{\alpha}_t}\mathbf{x}_0 + \sqrt{1 - \bar{\alpha}_t}\boldsymbol{\epsilon}_t, t)\|^2 \Big],
\label{eq:loss}
\end{equation}
where $\boldsymbol{\epsilon}_t$ is the Gaussian noise added at time $t$ in the forward pass, $\mathbf{x}_t$ is the corrupted data following \autoref{eq:diffusion}. In our case, $\mathbf{x}_0$ is the clean node features $F_p, F_z, E_p, E_{zv}$, and $\mathbf{x}_t$ is the noisy node features after forward pass. During sampling, the predicted noise at every time step is used to denoise the data from random Gaussian noise.

\subsection{B-rep Post-Processing}
\label{sec:post}
A set of heuristics is used to find duplicated nodes in the tree and explicitly decode the generated B-rep topology. For efficiency, we use early pruning to recover association as the tree is being generated. Given the face nodes, we detect duplicates as those with bounding box corners within 0.08 Euclidean distance of each other and averaged point-wise difference less than 0.2 for the decoded points from the latent shape feature. Duplicate faces are removed to recover the unique faces belonging to a solid. We avoid regenerating edges for the duplicated faces in the next stage. 

A similar procedure recovers the unique edges associated under every face, the same 0.08 edge bounding box threshold and 0.2 threshold for the decoded shape feature points are used to merge edges under the same parent. The vertex/edge association is always two and does not require deduplication. Finally, we traverse the tree again from leaf to root to find duplicated child nodes at the same level but associated with different parents, which are the shared edges and vertices for determining the final mating relations.

After recovering the mating and association relations, we perform an extra step of fine-tuning the geometry based on the generated topology. The vertex positions are averaged across its associated duplicates. Edge shape geometry points are scaled and translated to align with the associated start and end vertices. An edge is flipped if it is reversed with regard to its vertices. The face shape geometry points are also scaled and translated so that they tightly fit all the associated edges with the minimum Chamfer Distance.

\begin{figure}[h]
    \centering
    \includegraphics[width=0.99\columnwidth]{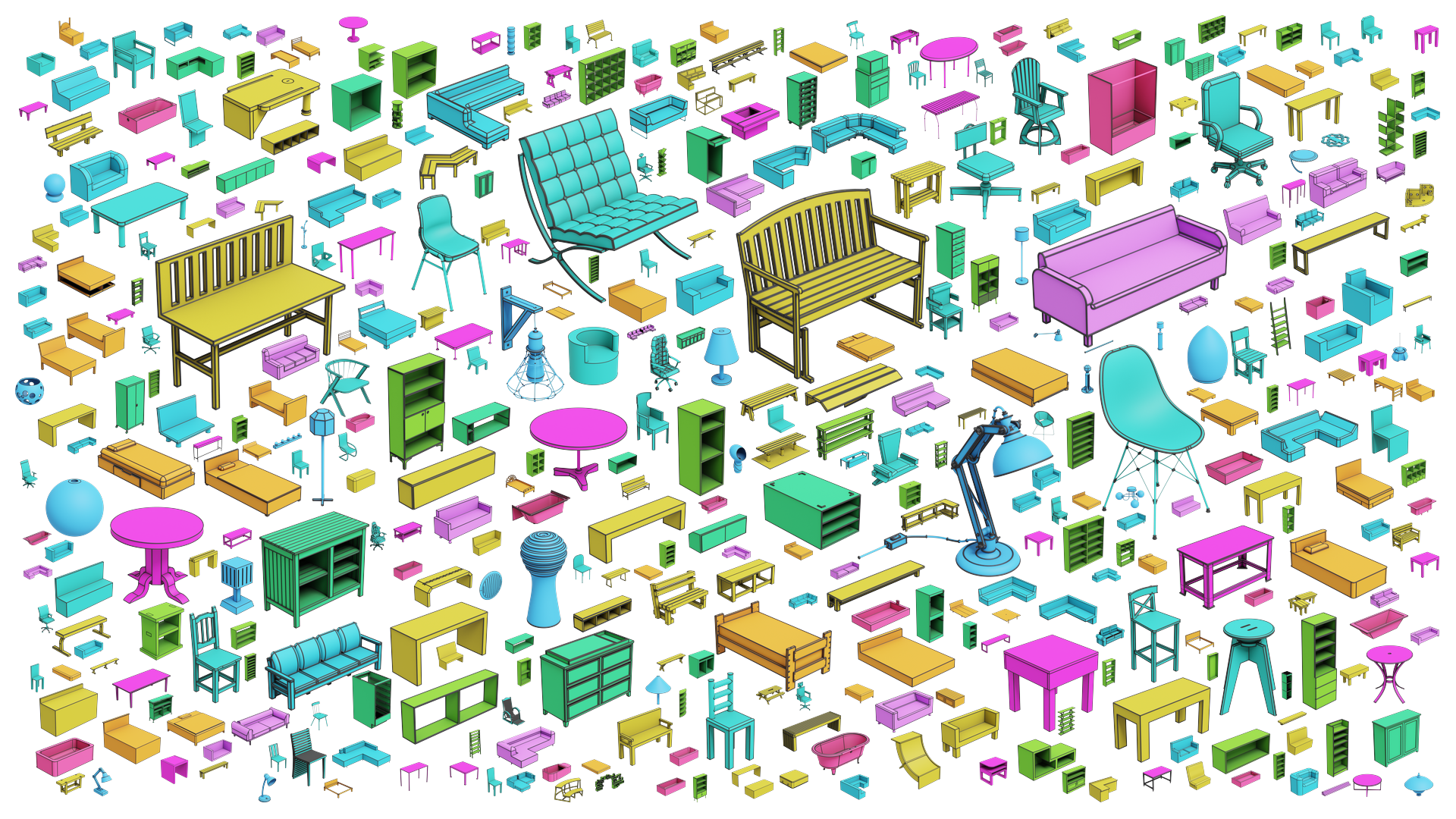}
    \caption{\datasetName\ overview colored by category.}
    \label{fig:dataset_overview}
\end{figure}

\section{Furniture B-Rep Dataset}
\label{sec:dataset}

To promote future B-rep generation research, we introduce the \datasetName\ containing 6,171 B-rep CAD models across 10 common furniture categories. To our knowledge, this is the first dataset in the B-rep format containing 3D models with freeform surfaces in addition to canonical category labels. 
%
We contacted PTC, the developers of Onshape~\cite{onshape}, a cloud-native CAD and PDM platform, and obtained an agreement to export publicly available design assets from the Onshape repository into our B-rep dataset. The Onshape public design library, consisting of millions of human-built models, is unique in the industry, providing an invaluable resource for CAD research and development.
We used keywords to identify 3D models in each category and searched inside both partstudio and assembly documents. We perform a number of manual filtering steps on the raw data to remove duplicates, verify correct categories, remove low quality data, and rotate the 3D models in each category into a canonical orientation. A visual overview of the CAD B-rep models in the \datasetName\ is provided in \autoref{fig:dataset_overview}. 

\begin{figure}[t]
    \centering
    \begin{subfigure}[b]{0.49\columnwidth}
        \centering
        \includegraphics[width=\textwidth]{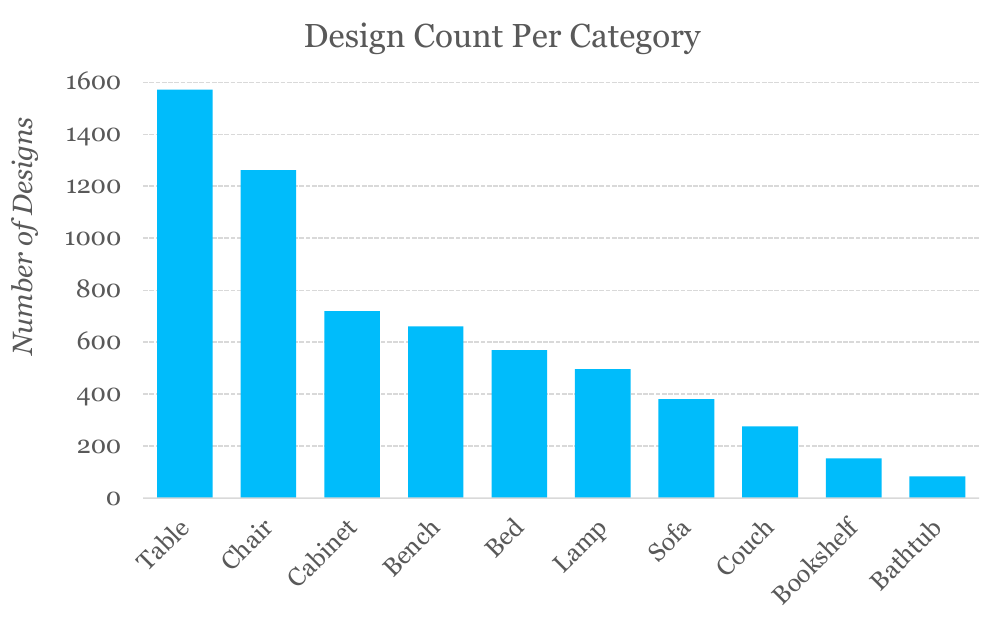}
        \caption[]%
        {{\small B-rep model count per category.}   } 
        \label{fig:dataset_classes}
    \end{subfigure}
    \hfill
    \begin{subfigure}[b]{0.49\columnwidth}  
        \centering 
        \includegraphics[width=\textwidth]{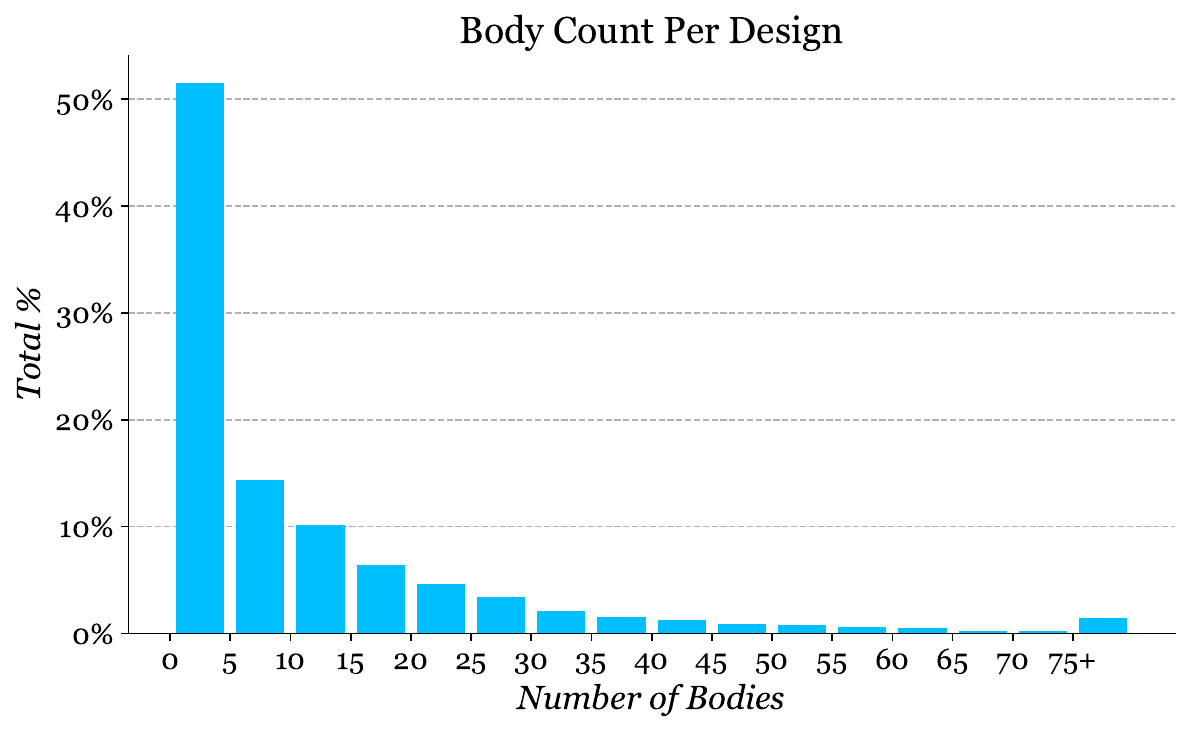}
        \caption[]%
        {{\small Number of solids per B-rep.}    }
        \label{fig:dataset_body_count}
    \end{subfigure}
    \par\bigskip 
    \vskip\baselineskip
    \begin{subfigure}[b]{0.49\columnwidth}   
        \centering 
        \includegraphics[width=\textwidth]{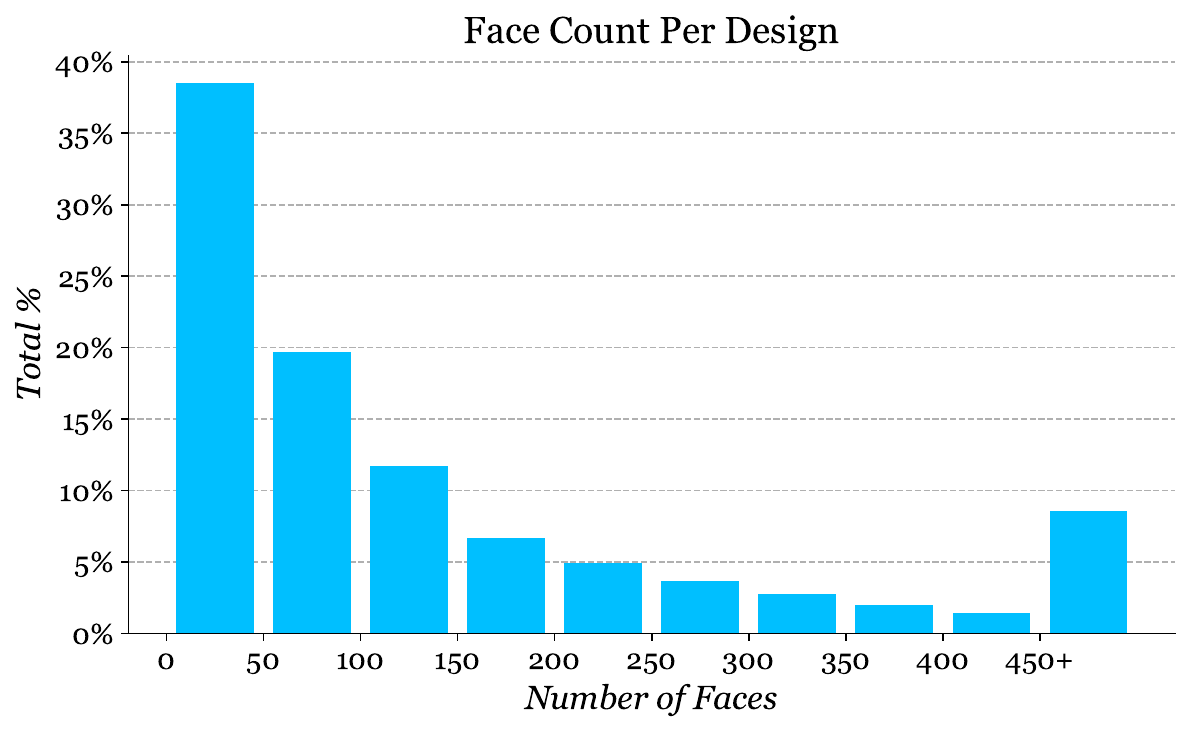}
        \caption[]%
        {{\small Number of faces per B-rep.}    }
        \label{fig:dataset_face_count}
    \end{subfigure}
    \hfill
    \begin{subfigure}[b]{0.49\columnwidth}   
        \centering 
        \includegraphics[width=\textwidth]{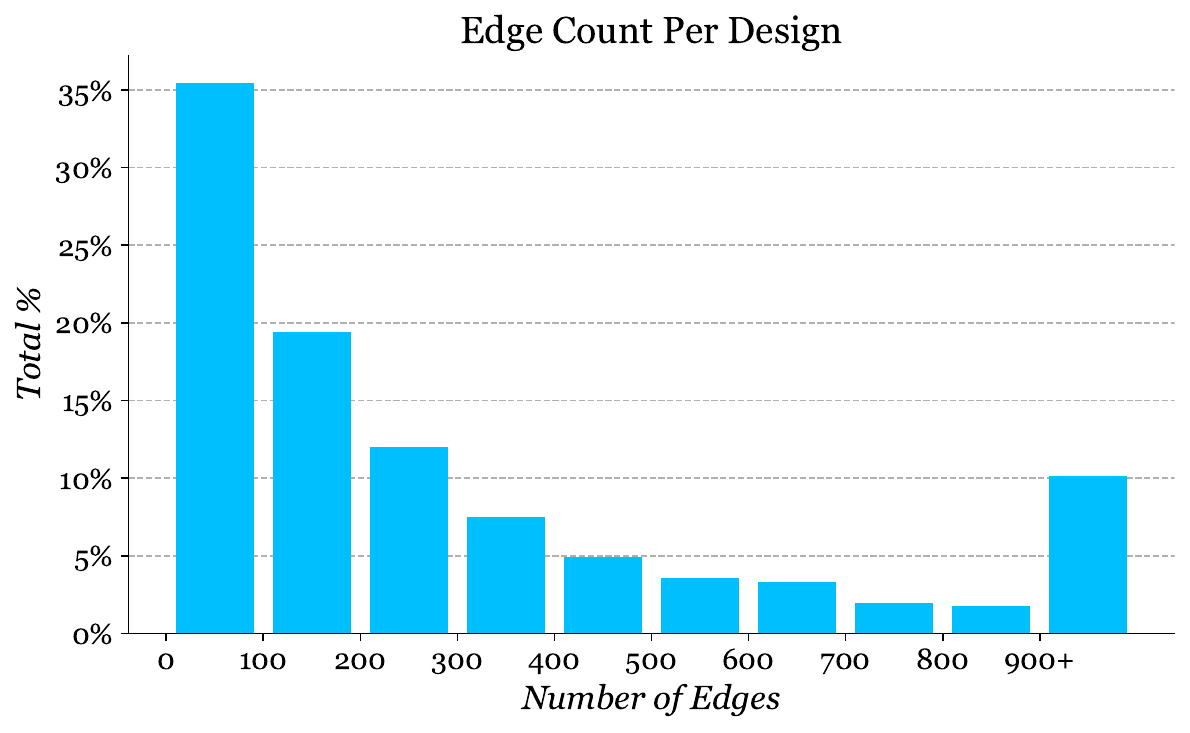}
        \caption[]%
        {{\small Number of edges per B-rep.}    }
        \label{fig:dataset_edge_count}
    \end{subfigure}
    \caption[ Statistics ]
    {\small Statistics for the \datasetName.}
    \label{fig:dataset_stats}
\end{figure}

Additional statistics for the \datasetName\ are provided in \autoref{fig:dataset_stats}.
\autoref{fig:dataset_classes} shows the category names in the dataset along with the number of B-rep models per category. 
\autoref{fig:dataset_body_count} provides the distribution for the number of solid bodies per B-rep in the dataset. Similarly, \autoref{fig:dataset_face_count} and \autoref{fig:dataset_edge_count} plot the distributions of the face and edge counts in every B-rep model.

\section{Experiments}
\label{sec:experiments}

This section presents unconditional and conditional generation results. Extensive analysis demonstrates that \modelName\ consistently produces high quality B-reps with complex topology and geometry, while providing enhanced control over the generation process.

\subsection{Experiment Setup}

\subsubsection{Datasets}
We evaluate generation performance on 1) the DeepCAD dataset~\cite{wu2021deepcad} of mechanical parts made from sketch and extrude operations, 
2) the \datasetName\ with more complicated furniture models, and 3) the ABC dataset \cite{koch2019:abc} containing a wide variety of parts from industrial designs. We use the original train/val/test split from DeepCAD and remove duplicated models in the training set following~\citep{willis2021engineering}. Closed faces (cylinders, etc.) are split on the seams following SolidGen~\cite{jayaraman2022solidgen}. B-reps with more than 30 faces or 20 edges per face, and made from multiple bodies are removed. After filtering, a total of 87,815 B-reps are used for training the VAEs and the latent diffusion module. We also randomly split the ABC and \datasetName\ into 90\%-5\%-5\% for train/val/test. In total, 259,597 ABC B-reps and 1,198 furniture B-reps with a maximum of 50 faces and 30 edges per face are used for training. Latent diffusion model is trained on all 10 categories for the \datasetName.

\begin{figure*}[ht]
    \begin{center}
    \includegraphics[width=0.99\textwidth]{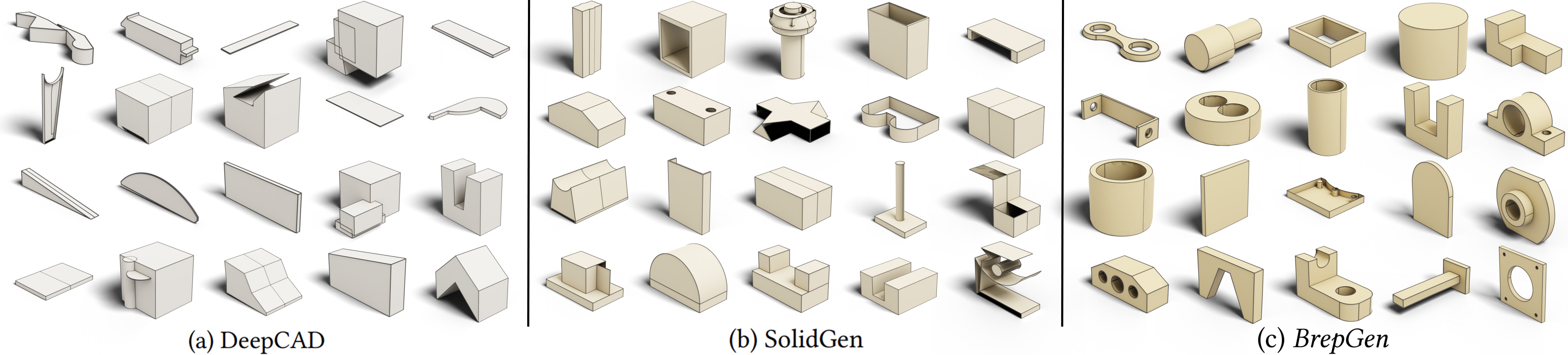}
        \caption{Unconditional generation results on DeepCAD mechanical parts by (a) DeepCAD, (b) SolidGen and (c) our method \modelName. Our method generates more realistic-looking CAD models with fewer broken geometry. Topological connections are also correct even on complicated objects. }
        \label{fig:uncond_deepcad}
    \end{center}
\end{figure*}

\subsubsection{Training} 
We implement \modelName\ in PyTorch and trained it on 4 NVIDIA RTX A5000 GPUs. Half-precision is used to speed up the training. We use AdamW \cite{loshchilov2018decoupled} with $5e-4$ learning rate. Gradient clipping of 5 are used for the VAE optimization. Weight for KL regularization is set to $1e-6$. Latent diffusion module uses 1,000 diffusion steps and a linear beta schedule from $1e-4$ to $0.02$.
The face and edge VAEs are trained for 400 epochs on DeepCAD (or 200 epochs on ABC) with a batch size of 512. Weights are further fine-tuned for 200 epochs for the \datasetName. The two face denoisers in the latent diffusion model are trained for 3,000 epochs with a batch size of 512, and the two edge denoisers are trained for 1,000 epochs with a batch size of 64. For the ABC dataset, we reduce the training epochs of the edge denoisers to 300, and the surface latent geometry denoiser to 1,000.

To alleviate train and test time difference, we perform cross-model augmentation where input to the conditional denoisers are randomly augmented following the scheme used in \cite{ho2022cascaded}. Association duplication randomly selects and pads the edges and faces in a B-rep (with repetition) until reaching the predefined maximum length, which is 30/600 for the face/edge denoisers of DeepCAD, and 50/1500 for the face/edge denoisers of the Furniture B-Rep and ABC dataset. All bounding boxes are normalized to [-3, 3] range.

\subsubsection{Inference}
At inference time, we use PNDM~\cite{liu2022pseudo} of 200 forward passes for fast sampling, with the exception of the face and edge position denoisers where we switch to a slower DDPM without strided sampling from $T=250$ to $0$. Empirically, this coarse-to-fine denoising generates a more precise bounding box position. When sampling from random Gaussian noise, increasing the number of tokens beyond the maximum threshold used in training is helpful. This avoids noisy nodes being incorrectly merged early in the denoising process, leading to a lack of available primitives to fill in missing regions. For the DeepCAD dataset we denoise B-reps with a maximum of 50 faces and 30 edges per face. For the \datasetName\ and ABC, 80 faces and 40 edges per face are denoised. After denoising and post-processing, we use the OpenCascade functions \textit{GeomAPI\_PointsToBSplineSurface} and \textit{GeomAPI\_PointsToBSpline} to approximate the face and edge points with B-Spline surfaces and curves. The connected loops bound the surface, and the trimmed faces are sewn together to form the final B-rep solid. On a RTX A5000 GPU,  the four denoisers in total require an average of around 5 seconds to generate a B-rep trained on DeepCAD data, and around 10 seconds for the more complicated Furniture and ABC data.

\subsubsection{Network Architecture} The face VAE network has four downsampling and four upsampling blocks. Each block contains two layers of 2D-ResNet blocks with skip connections. The feature dimensions of the downsampling and upsampling blocks are  512-512-256-128 and 128-256-512-512, respectively. The edge VAE has similar ResNet blocks but uses 1D convolution instead. The diffusion model is a standard Transformer module with pre-layer normalization, 12 self-attention layers with 12 heads, a hidden dimension of 1024, feature dimension of 768, and a 0.1 dropout rate.

\subsubsection{Evaluation Metrics}
We use two sets of metrics to quantitatively measure generation quality: \textit{Distribution Metrics} and \textit{CAD Metrics}. For \textit{Distribution Metrics} we use 3,000 B-reps randomly-sampled from the generated data and 1,000 B-reps from the reference test set. For each B-rep, we sample 2,000 points from the solid surface and compute the following metrics:

\noindent $\bullet$ \textit{Coverage} (COV) is the percentage of reference data with at least one match after assigning every generated data to its closest neighbor in the reference set based on Chamfer Distance (CD). 

\noindent $\bullet$ \textit{Minimum Matching Distance} (MMD) is the averaged CD between a reference set data and its nearest neighbor in the generated data.

\noindent $\bullet$ \textit{Jensen-Shannon Divergence} (JSD) measures the distribution distance between reference and generated data after converting point clouds into $28^3$ discrete voxels.

For \textit{CAD Metrics} the same 3,000 B-reps are used to compute the following metrics, as used by~\cite{jayaraman2022solidgen}:

\noindent $\bullet$ \textit{Novel} percentage of data that do not appear in training set. 

\noindent $\bullet$ \textit{Unique} percentage of data that appears only once in generation. 

\noindent $\bullet$ \textit{Valid} percentage of B-rep data that are watertight solids.

\begin{figure*}[ht]
    \begin{center}
    \includegraphics[width=0.99\textwidth]{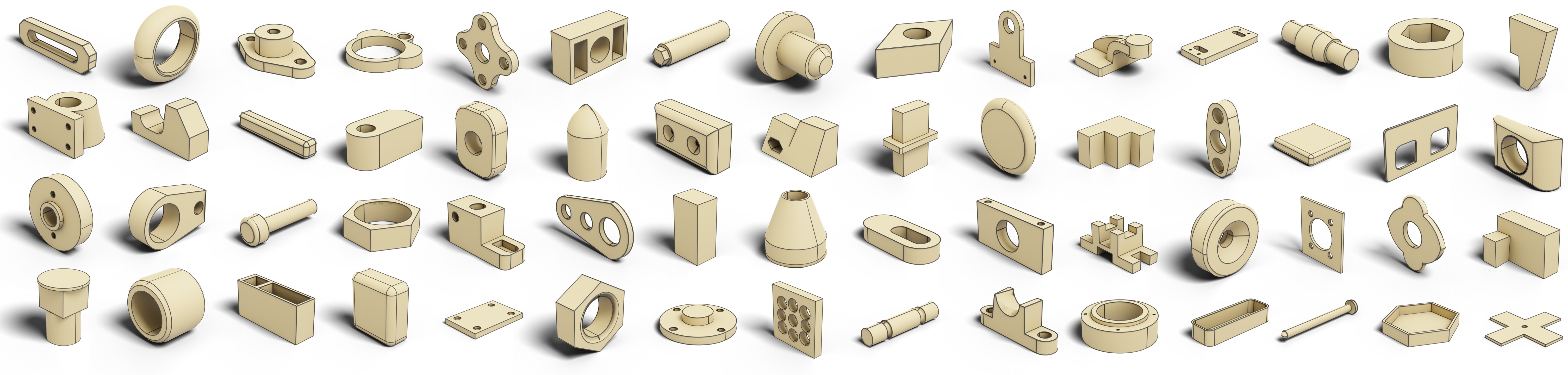}
        \caption{Unconditional generation results on the ABC dataset. \modelName~generates realistic CAD B-reps with complicated geometry and topology. }
        \label{fig:uncond_abc}
    \end{center}
\end{figure*}

\subsection{Unconditional B-rep Generation}
We compare with DeepCAD \cite{wu2021deepcad} and SolidGen \cite{jayaraman2022solidgen} for unconditional generation of mechanical parts. DeepCAD is evaluated based on the reconstructed B-reps from its generated sketch and extrude sequences. We also evaluate \modelName\ on the large-scale ABC dataset~\cite{koch2019:abc}.

\begin{table}
\caption{Quantitative evaluations for DeepCAD unconditional generation based on the \textit{Coverage} (COV) percentage, \textit{Minimum Matching Distance} (MMD),  \textit{Jensen-Shannon Divergence} (JSD), and \textit{Unique}, \textit{Novel}, \textit{Valid} ratio. The bottom row reports \modelName\ results on the ABC dataset. Both MMD and JSD are multiplied by $10^2$
.}
\label{tab:deepcad_uncond}
\begin{center}
\setlength\tabcolsep{3.5 pt}
\small
\begin{tabular}{lccc|ccc}
\toprule
       Method  & COV   & MMD  & JSD  & Novel   & Unique & Valid    \\
         &  \% $\uparrow$  & $\downarrow$  & $\downarrow$  &\% $\uparrow$   & \% $\uparrow$  & \% $\uparrow$   \\ \midrule        
DeepCAD        &   65.46     &  1.29    &   1.67    & 87.4 & 89.3 &  46.1\\
SolidGen       &   71.03     &  1.08    &   1.31    & 99.1 & 96.2 & 60.3\\
\modelName\    &   73.87     &  1.04    &   1.28    & 99.8 & 99.7 & 62.9\\ \midrule   
\modelName\ (ABC)    &   57.92      &  1.35     &  3.69    & 99.7 & 99.4 & 48.2\\ 
\bottomrule
\end{tabular}
\end{center}
\end{table}

\subsubsection{Quantitative Evaluation}
The first three rows in \autoref{tab:deepcad_uncond} report the averaged scores on DeepCAD data from 20 different runs. \modelName\ consistently outperforms both baselines with better COV, MMD scores and a substantially lower JSD; demonstrating improvements in generation quality and a closer match to the ground-truth distribution. Generated data from \modelName\ is also novel and different from the training set, as indicated by the high novel and unique scores. Here, we consider two B-reps identical if the topology connections are the same, and the shape geometry points from the faces are equal after 4 bit quantization. The valid ratio of \modelName\ is also better compared to the SolidGen direct B-rep generation baseline. Our criteria for validity enforces watertight CAD models with no broken topology. Novel and unique scores are calculated using the valid B-reps. Last row in \autoref{tab:deepcad_uncond} also reports the \modelName\ quantitative results for ABC unconditional generation.

\subsubsection{Qualitative Evaluation}
\begin{figure}
    \centering
    \includegraphics[width=\columnwidth]{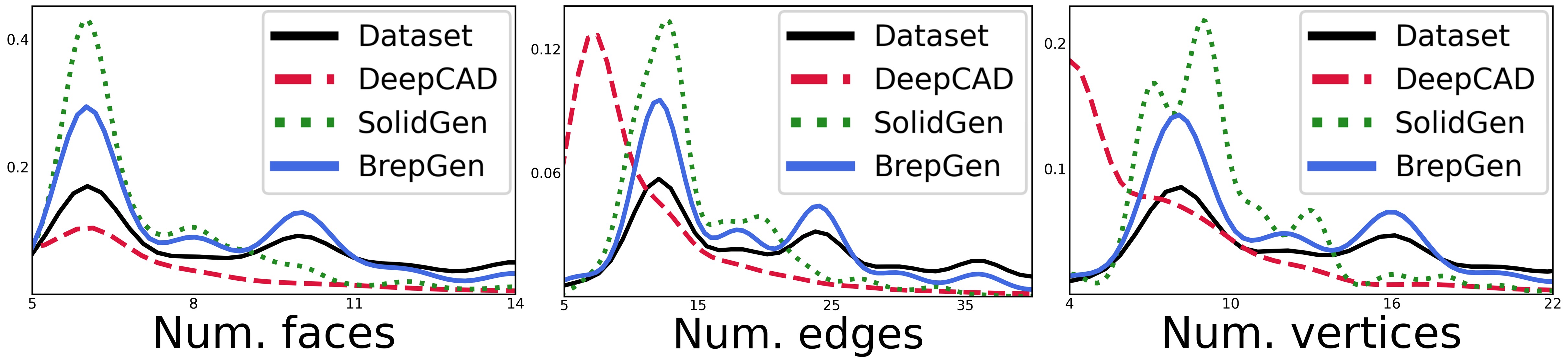}
    \caption{Generated B-rep data distributions, for the number of faces, edges, and vertices, compared with the DeepCAD training set and other baselines.}
    \label{fig:deepcad_dist}
\end{figure}

\autoref{fig:uncond_deepcad} illustrates qualitative results for unconditional DeepCAD generation. We see that \modelName\ generates appealing B-reps made of a wide range of topologically connected faces, showing better generation diversity than the baselines. The structural complexity is also higher with fewer unbounded open regions or self-intersecting edges. Additional results are provided in \autoref{fig:deepcad_extra}. \autoref{fig:uncond_abc} also shows the unconditional ABC generation results by \modelName. We observe that our method effectively trains on the large-scale ABC dataset and generates realistic, diverse CAD B-reps of complicated topology and geometry. \autoref{fig:deepcad_dist} further compares the distribution of B-reps generated using \modelName\ for the number of faces, edges, and vertices, with the training set and other baselines. The plot shows that our method has the best matching curves to the ground-truth DeepCAD training data.

\begin{figure}
    \centering
    \includegraphics[width=0.98\columnwidth]{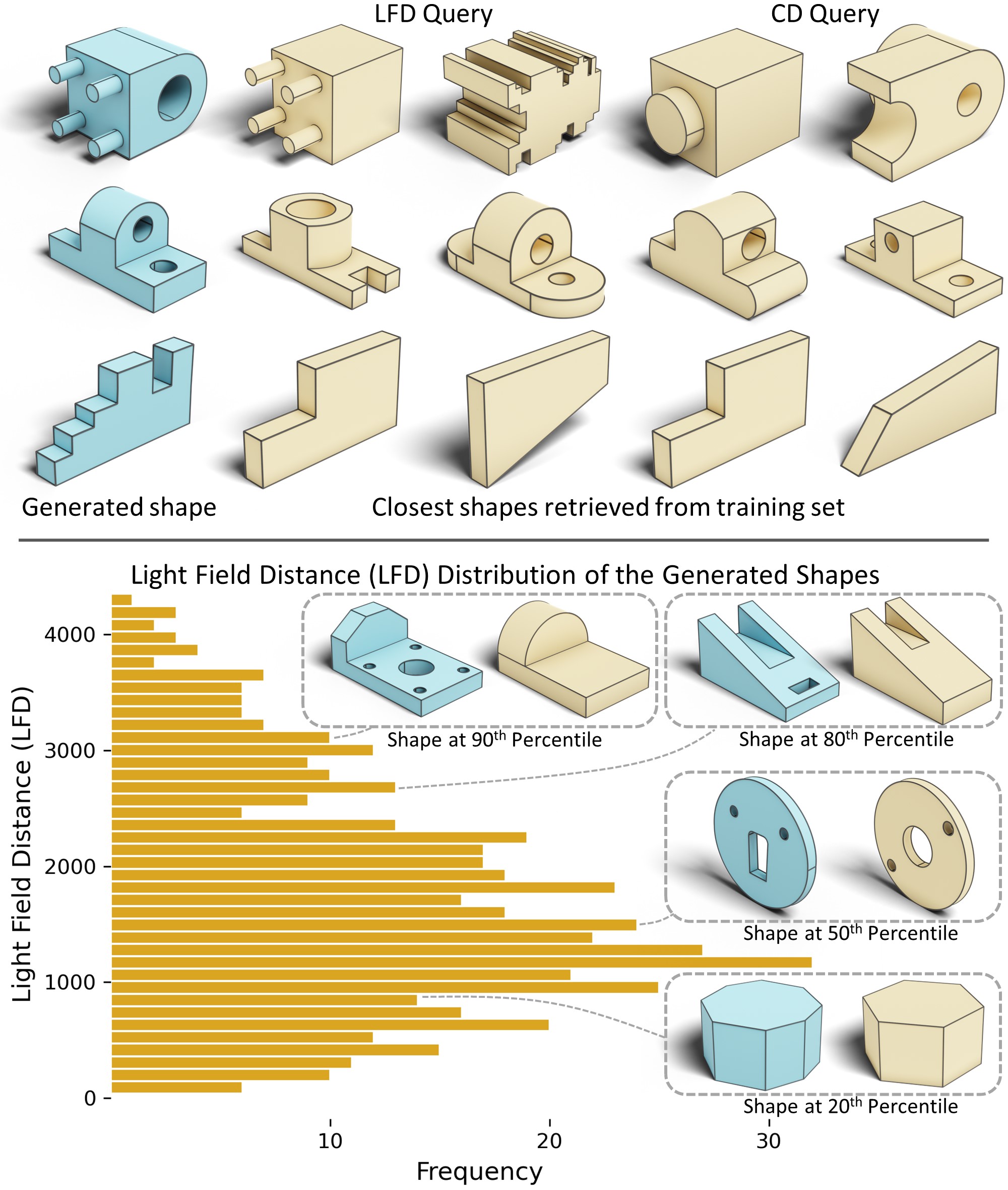}
    \caption{Novelty analysis of the generated DeepCAD shapes. Top: Given generated shapes (blue), we visualize the top-two most similar shapes (yellow) retrieved from the training set using LFD or CD. Bottom: LFD distribution between 500 randomly generated data and their most-similar retrieved training shapes. Visual results at different percentiles are shown, where generated shapes can be both realistic and novel.}
    \label{fig:nns}
\end{figure}

\begin{figure*}[t]
    \begin{center}
    \includegraphics[width=0.95\textwidth]{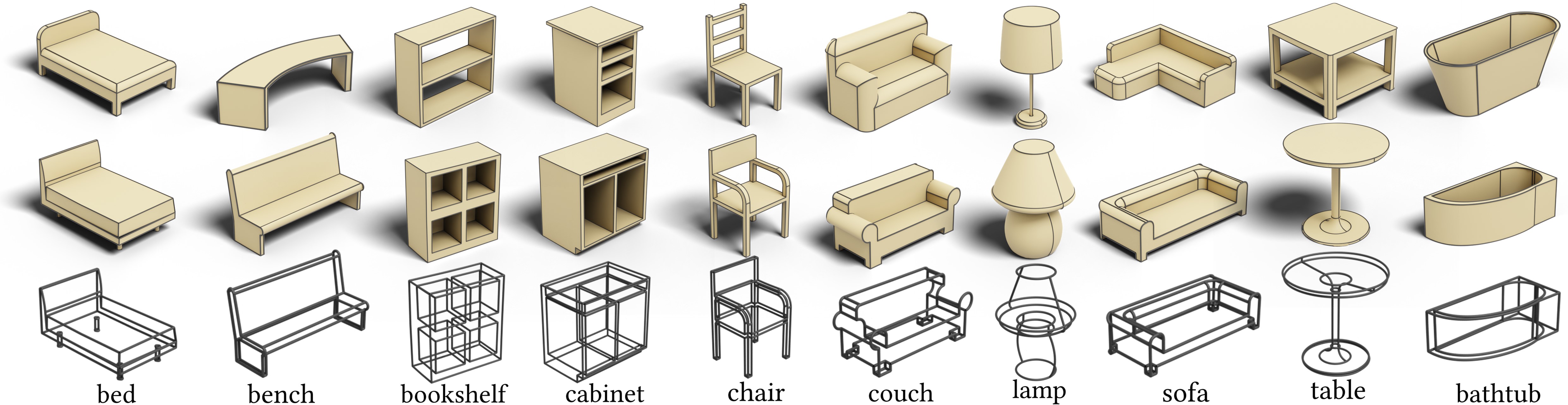}
        \caption{\modelName\ generation results conditioned on the class label. Last row shows the wireframe connected by the generated edges and vertices. The results clearly show that \modelName\ excels in generating well-structured CAD models with free-form surfaces and curvatures.}
        \label{fig:cond_furniture}
    \end{center}
\end{figure*}

\subsubsection{Shape Novelty Analysis}
To further evaluate the ability of \modelName\ to generate novel shapes, we adopt the settings from Neural Wavelet~\cite{hui2022neural} and retrieve the most similar shapes from the DeepCAD training set using Chamfer Distance (CD) and Light Field Distance (LFD)~\cite{chen2003visual}. \autoref{fig:nns} (top) shows the generated shapes (blue), together with the top-two most similar training shapes (yellow) retrieved using LFD or CD. Additionally, \autoref{fig:nns} (bottom) plots the LFD distribution between 500 randomly generated samples and their closest retrieved results from the training set. Results show that \modelName\ generates realistic and visually accurate CAD models, while featuring novel topology and geometry different from the training data. 

\begin{figure}
    \centering
    \includegraphics[width=0.96\columnwidth]{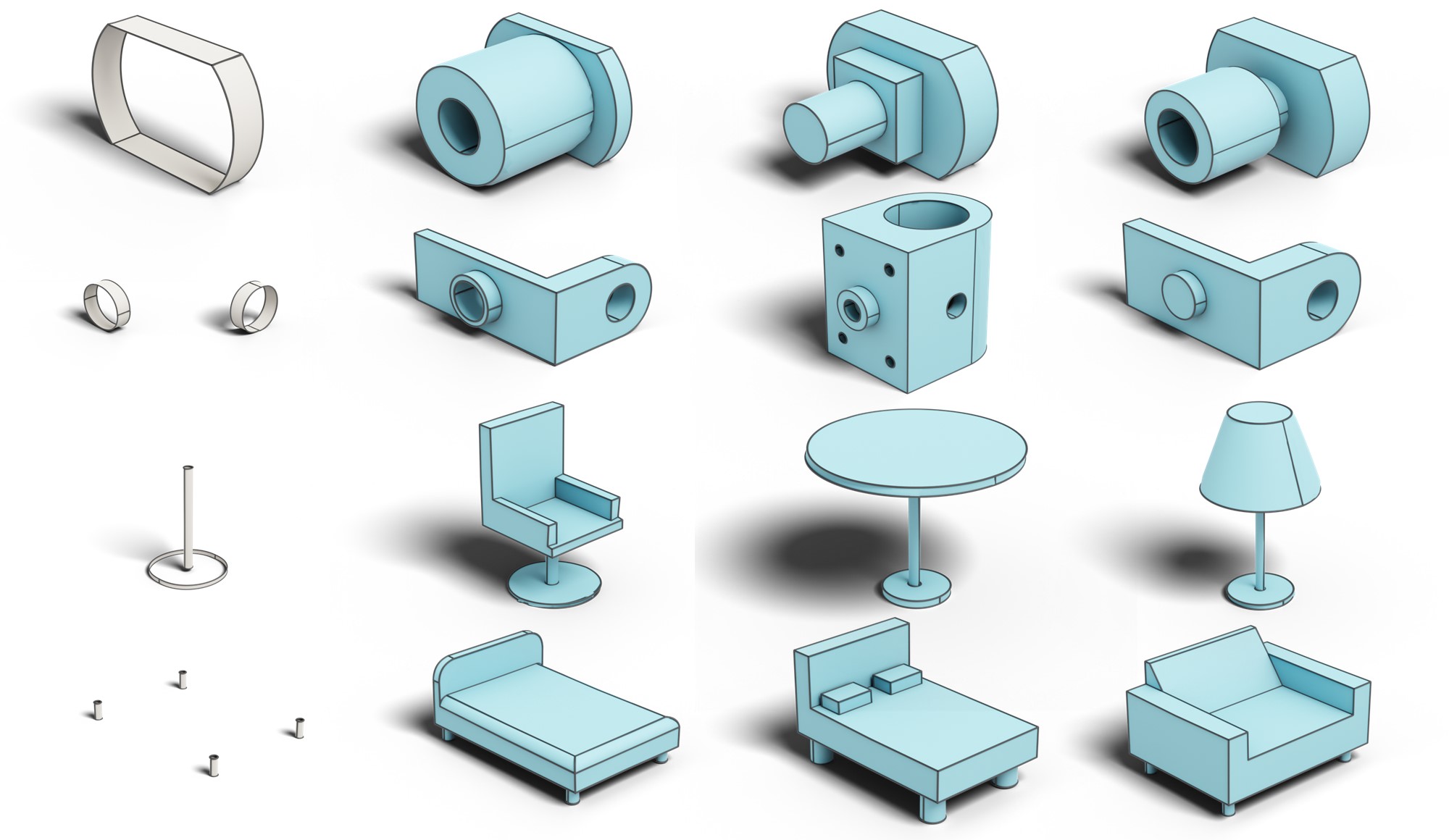}
    \caption{Given partial faces (gray), \modelName\ auto-completes the full B-rep geometry and topology, generating a diverse set of CAD models (blue).}
    \label{fig:inpaint}
\end{figure}

\subsection{Controllable B-rep Generation}
We demonstrate \modelName\ results on \datasetName\ and show two applications for CAD design. With no positional encoding and random shuffling at training, \modelName\ is permutation invariant with respect to the input tokens, eliminating the need for a specific sequence order between existing and to-be-generated components.

\subsubsection{Class-conditioned Generation} A class-conditioned \modelName\ is trained on the \datasetName\ using classifier-free guidance \cite{ho2022classifier}. Class embeddings are added to every input token and an extra label indicates the unconditional case, which occurs with a 10\% probability at training. \autoref{fig:cond_furniture} shows the generation results for all 10 categories. \modelName\ generates the correct topological connections across different components and outputs free-form surfaces not possible with SolidGen. The generated edges form connected wireframes with loops precisely trimming the faces into water-tight solids. Additional results are provided in \autoref{fig:furniture_extra}.

\subsubsection{CAD Autocompletion}
We use a pretrained \modelName\ to autocomplete the full B-rep from partial faces provided by the user. Motivated by RePaint~\cite{lugmayr2022repaint}, a random subset of face tokens are replaced with the provided face geometry diffused to that time step during face denoising. The replacement is discontinued in the final 50 steps to allow minor adjustments to the provided faces. Subsequent edges and vertices are also regenerated. \autoref{fig:inpaint} shows diverse autocompleted results from different Gaussian noise. Disjoint parts are connected and transformed into water-tight solids with faces seamlessly generated between, below, or above the given input. The inpainting process can make slight adjustments to user input so that it connects with the rest of the geometry. For example, the radius of the legs slightly increase for the bed in the last row.

\begin{figure}[h]
    \centering
    \includegraphics[width=0.97\columnwidth]{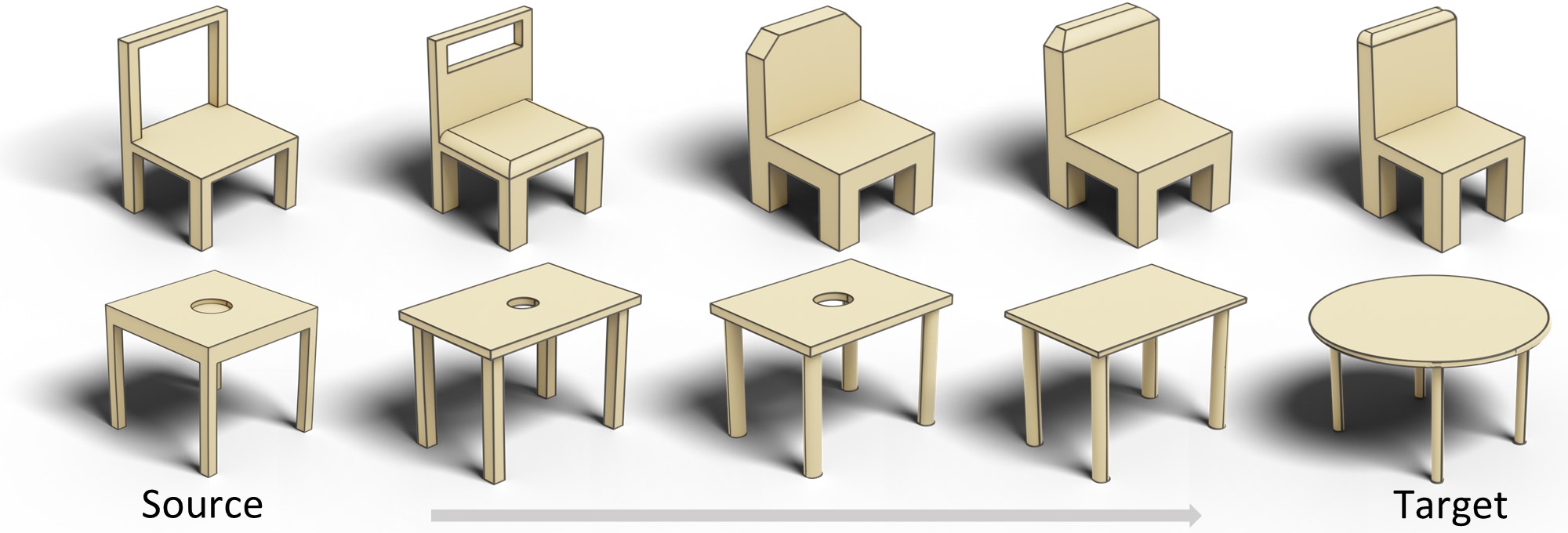}
    \caption{Design interpolation from source to target. The geometry and topology of the generated B-rep undergo progressive changes.}
    \label{fig:interpolate}
\end{figure}

\subsubsection{Design Interpolation}
Given a pair of CAD models, face tokens from the two shapes are concatenated, diffused for 150 steps, repeatedly padded and passed to the face denoiser. Edges and vertices are also regenerated. This is different from blending the latent value as mixing two geometry tokens will not lead to valid shapes. \autoref{fig:interpolate} shows the interpolated results. From left to right, face tokens from the target are increasingly added to the source for face generation, after which the source tokens are removed until only the target tokens are left. The topology and geometry changes are smooth, with interpolated design directly generated in B-rep format.

\begin{figure}
    \centering
    \includegraphics[width=0.9\columnwidth]{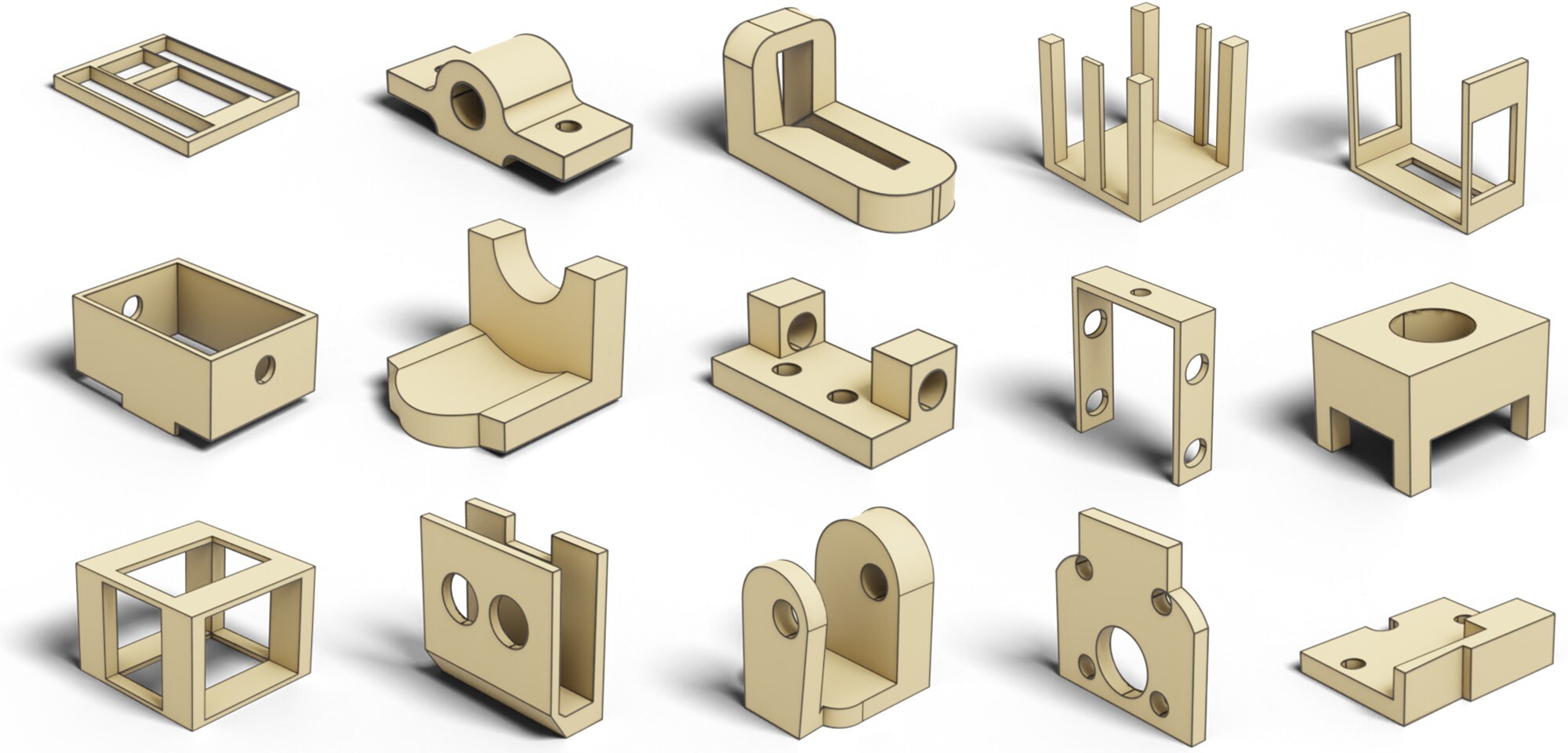}
    \caption{DeepCAD generation results conditioned on ground-truth topology.}
    \label{fig:gt_topo_brepgen}
\end{figure}

\begin{table}
\caption{Quantitative evaluations for DeepCAD generation with ground-truth topology shown as \modelName*. Second row reports the unconditional topology generation using multinomial diffusion. Both MMD and JSD are multiplied by $10^2$.}
\label{tab:abal_studies}
\begin{center}
\setlength\tabcolsep{3.5 pt}
\small
\begin{tabular}{lccc|ccc}
\toprule
       Method  & COV   & MMD  & JSD  & Novel   & Unique & Valid    \\
         &  \% $\uparrow$  & $\downarrow$  & $\downarrow$  &\% $\uparrow$   & \% $\uparrow$  & \% $\uparrow$   \\ \midrule        
\modelName*   &   78.16     &  1.02    &   0.90    & 99.9 & 97.6 & 79.8\\
Topology    &   -     &  -    &   -   & 95.1 & 64.5 & 6.2  \\
\bottomrule
\end{tabular}
\end{center}
\end{table}

\subsection{Ablation Studies} 

\subsubsection{Two-stage Generation}
To demonstrate the effectiveness of our unified generation approach, we conduct an ablation study where topology and geometry are separately generated in two stages. 
In the second stage, we modify \modelName\ to generate geometry conditioned on the given ground-truth topology. The node duplication process is removed and replaced by the graph attention used in HouseDiffusion~\cite{shabani2023housediffusion} to encode the topology relations. The first row in \autoref{tab:abal_studies} shows the conditional generation results trained on DeepCAD data, with visual results shown in \autoref{fig:gt_topo_brepgen}. Noticeably, the coverage and valid ratios increase, and the generated B-reps exhibit more intricate structures. We attribute this improvement primarily to knowing the ground-truth topology.

For the initial topology stage, we design a multinomial diffusion baseline \cite{hoogeboom2021argmax} that unconditionally generates the face-edge, edge-vertex incidence matrix as a $128 \times 128$ binary image. We observe that it is challenging to unconditionally generate the correct incidence matrix. Last row in \autoref{tab:abal_studies} reports a very low valid ratio. Topology is deemed valid only if every edge is connected to two faces and two vertices, and edges on each face form closed loops. Broken topology makes it difficult to integrate the two modules and output valid B-reps. This further shows the advantage of our structured latent geometry, which unifies topology and geometry and jointly generates rather than in two separate stages.

\subsubsection{Post-process Thresholds}
\begin{figure}
    \centering
    \includegraphics[width=0.7\columnwidth]{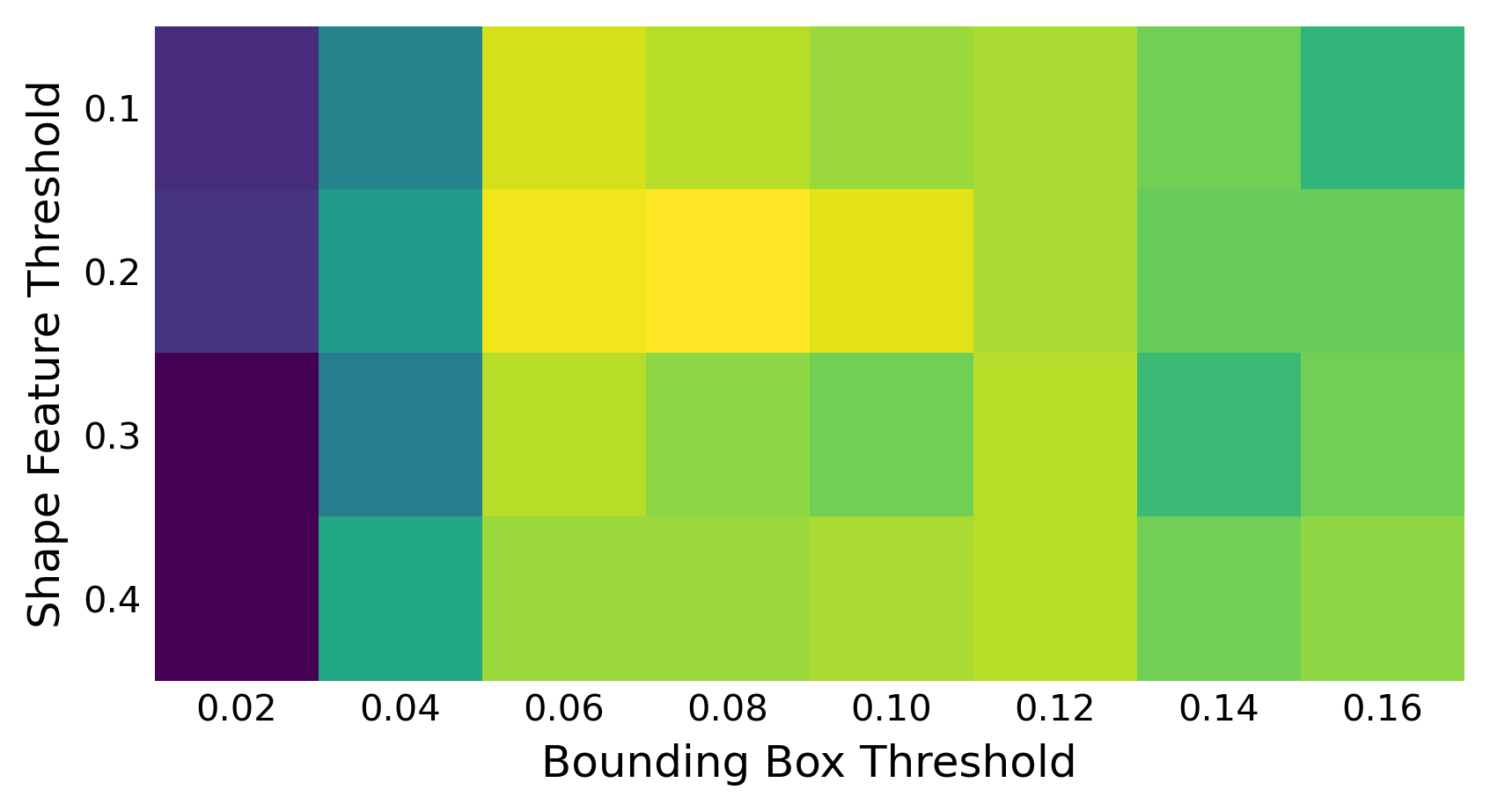}
    \caption{Valid ratio under different threshold values. Brighter color indicates higher ratio value. Result are averaged over 100 randomly generated samples from a model trained on the DeepCAD validation set.}
    \label{fig:threshold_abla}
\end{figure}

To study the impact of different thresholds on B-rep post-processing, we perform a grid search of the bounding box and decoded shape feature threshold values from \autoref{sec:post}. Concretely, we trained a latent diffusion model on the DeepCAD validation set and report the ratio of valid B-reps that are watertight from 100 randomly generated samples. \autoref{fig:threshold_abla} shows the valid ratios obtained under various threshold combinations. From the figure, we see that the best ratio is achieved when the bounding box threshold is within the range of $0.06$ to $0.1$, and the shape feature threshold is around $0.2$. A large bounding box threshold tends to incorrectly merge unique edges or faces, whereas a very small threshold fails to effectively identify duplicated geometry.

\subsection{Failure Cases} 
\begin{figure}[h]
    \centering
    \includegraphics[width=0.92\columnwidth]{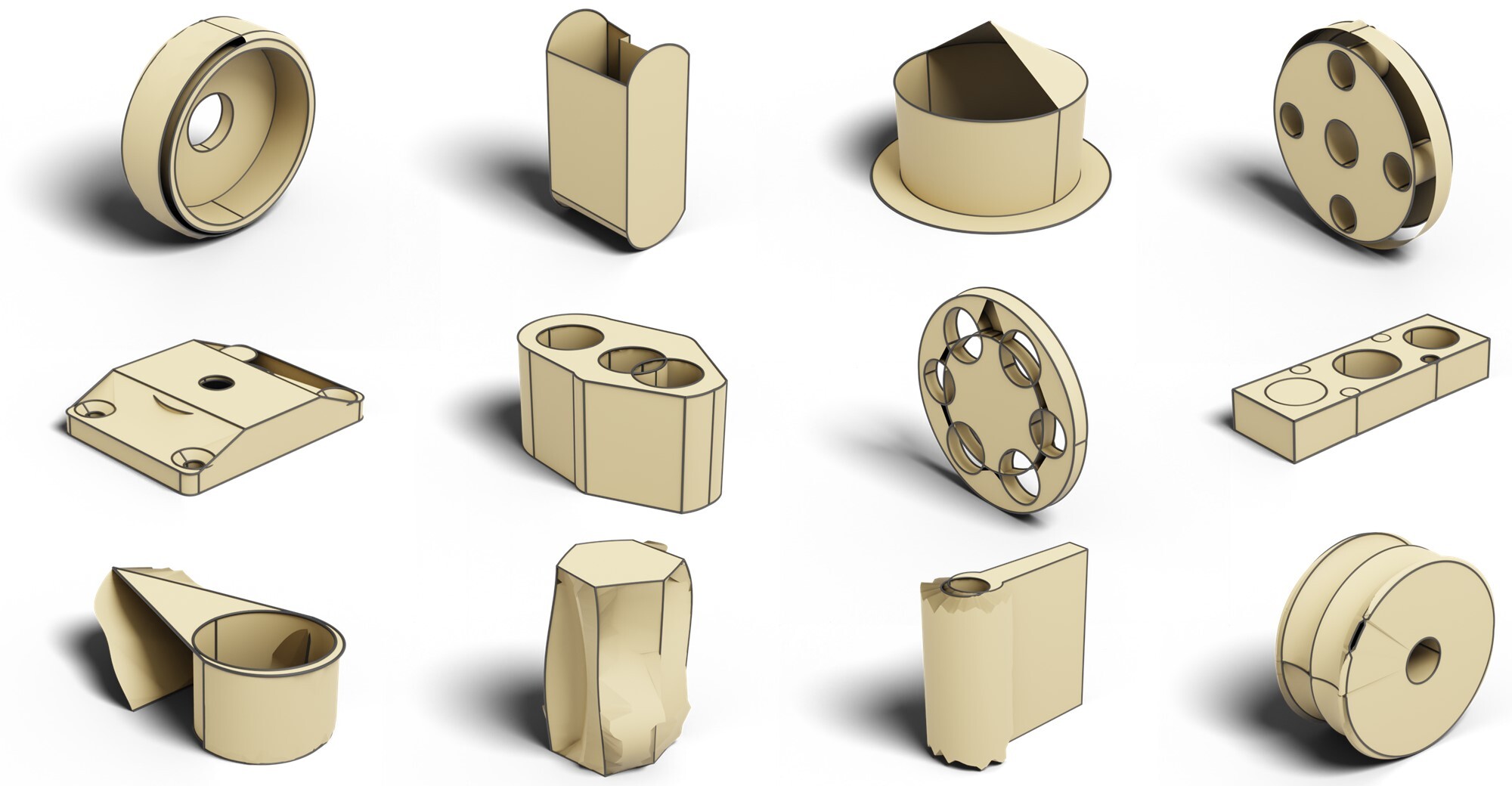}
    \caption{Three common failure cases categories for \modelName. Top row: non-watertight solids. Middle row: self-intersecting edges or faces. Bottom row: wobbly and broken geometry. }
    \label{fig:fail}
\end{figure}

\autoref{fig:fail} shows B-reps generated by \modelName\ from three common failure case categories. 1) Missing faces resulting in non-watertight solids (top row).  This failure is anticipated given \modelName\ can not guarantee watertight outputs. A potential solution involves detecting and denoising additional faces in the open regions. 2) Self-intersection of edges or faces leading to broken geometry after trimming (middle row). Future integration of a self-intersection loss might reduce such occurrences. 3) Wobbly and broken geometry (bottom row). This is mostly caused by noise in the decoded points, or inconsistent geometry between a surface and its connected edges.

\section{Limitations and Future Work}
\label{sec:limitations}
\modelName\ enables the generation of 3D models in the Boundary representation but has several limitations that warrant future work. \modelName\ supports only single body solids; more complicated CAD models with multiple assembled bodies are left to future work. If edges or faces are too close to one another they will be merged and identified as one after deduplication. In practice, this implies a minimum threshold, determining how close two similar surfaces or edges can be. We train on data with a minimum threshold of $0.05$ after normalizing the CAD geometry to the $[-3, 3]$ range. This limitation is similar to other sketch-and-extrude generation approaches, e.g. \cite{xu2022skexgen}, that quantize geometry to a set bit range resulting in vertex merging. The chosen threshold of $0.05$ is roughly equivalent to 1 bin difference after 7 bit quantization. \modelName\ does not guarantee  watertight solids.  Small gaps may exist due to the slow convergence of the denoising process.  Occasionally entire faces are also missing. Experimentally we find that the valid ratio for \modelName, which includes watertightness, is higher than for other architectures. Finally, while the heuristic post-processing used to generate the B-rep is simple, fast, and can handle complicated data, to achieve better results future work on a learning-based post-processing module may provide more robust handling of invalid shapes.

\section{Conclusion}
\label{sec:conclude}

We introduced \modelName, a generative diffusion model for direct B-rep generation. Extensive experiments showed that \modelName\ surpasses existing methods and establishes state-of-the-art results for B-rep generation. We outlined our newly collected \datasetName\ and results that demonstrate the ability of  \modelName\ to generate complicated B-rep 3D models with free-form and doubly-curved surfaces for the first time. We believe \modelName\ moves us one step closer towards an automatic system capable of directly generating B-reps to reduce the extensive manual labor required from skilled designers using today's professional CAD software.


\begin{acks}

This research is partially supported by NSERC Discovery
Grants with Accelerator Supplements and DND/NSERC
Discovery Grant Supplement, NSERC Alliance Grants, and
John R. Evans Leaders Fund (JELF). 
We also extend our thanks to PTC for their support and providing access to Onshape, whose extensive library of publicly available, human-built CAD models is instrumental to this work.
\end{acks}

\bibliographystyle{ACM-Reference-Format}
\bibliography{bibliography}

\newpage
\appendix

\begin{figure*}
    \begin{center}
    \includegraphics[width=0.9\textwidth]{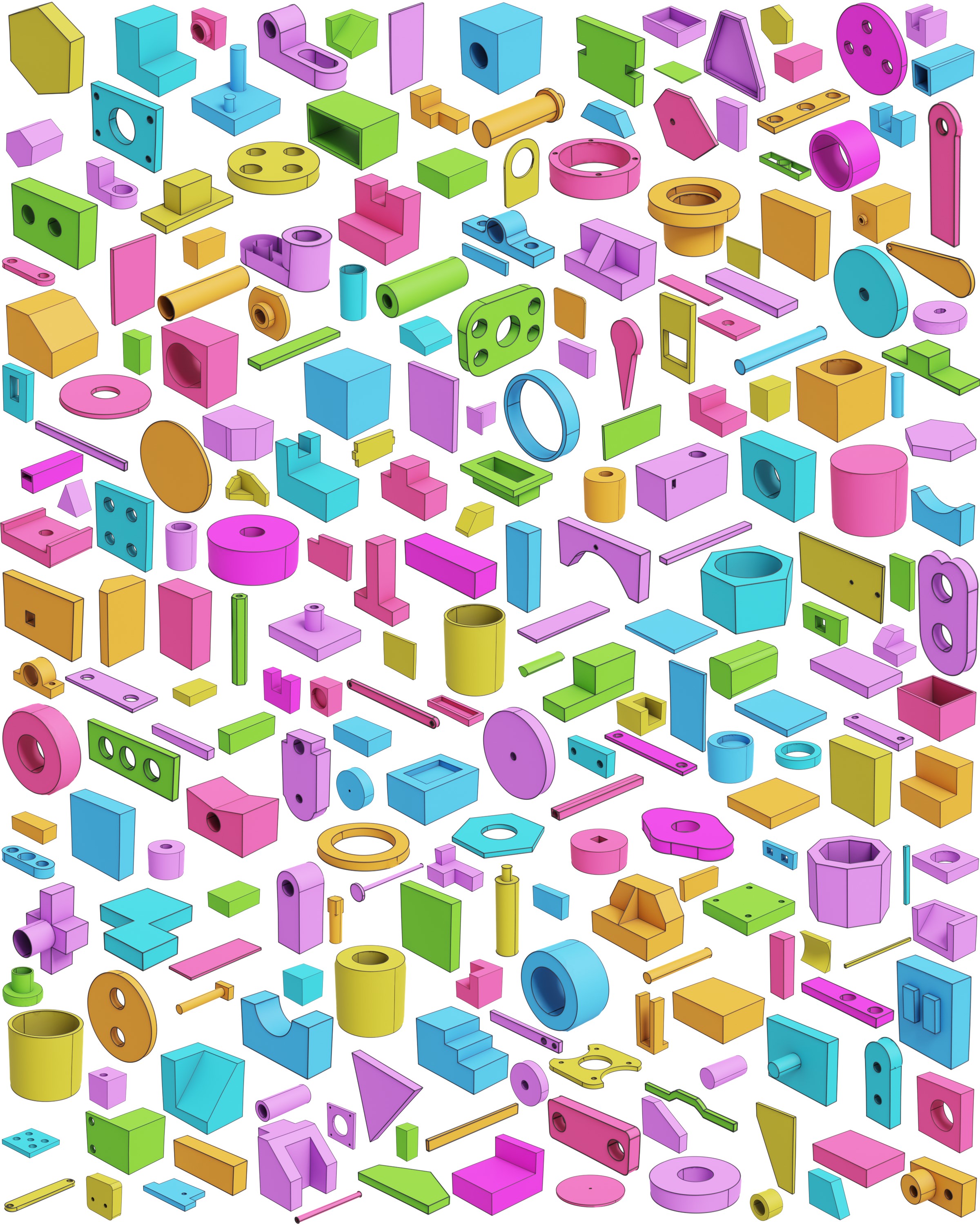}
        \caption{Unconditionally generated mechanical part B-reps from \modelName.}
        \label{fig:deepcad_extra}
    \end{center}
\end{figure*}

\begin{figure*}
    \begin{center}
    \includegraphics[width=0.9\textwidth]{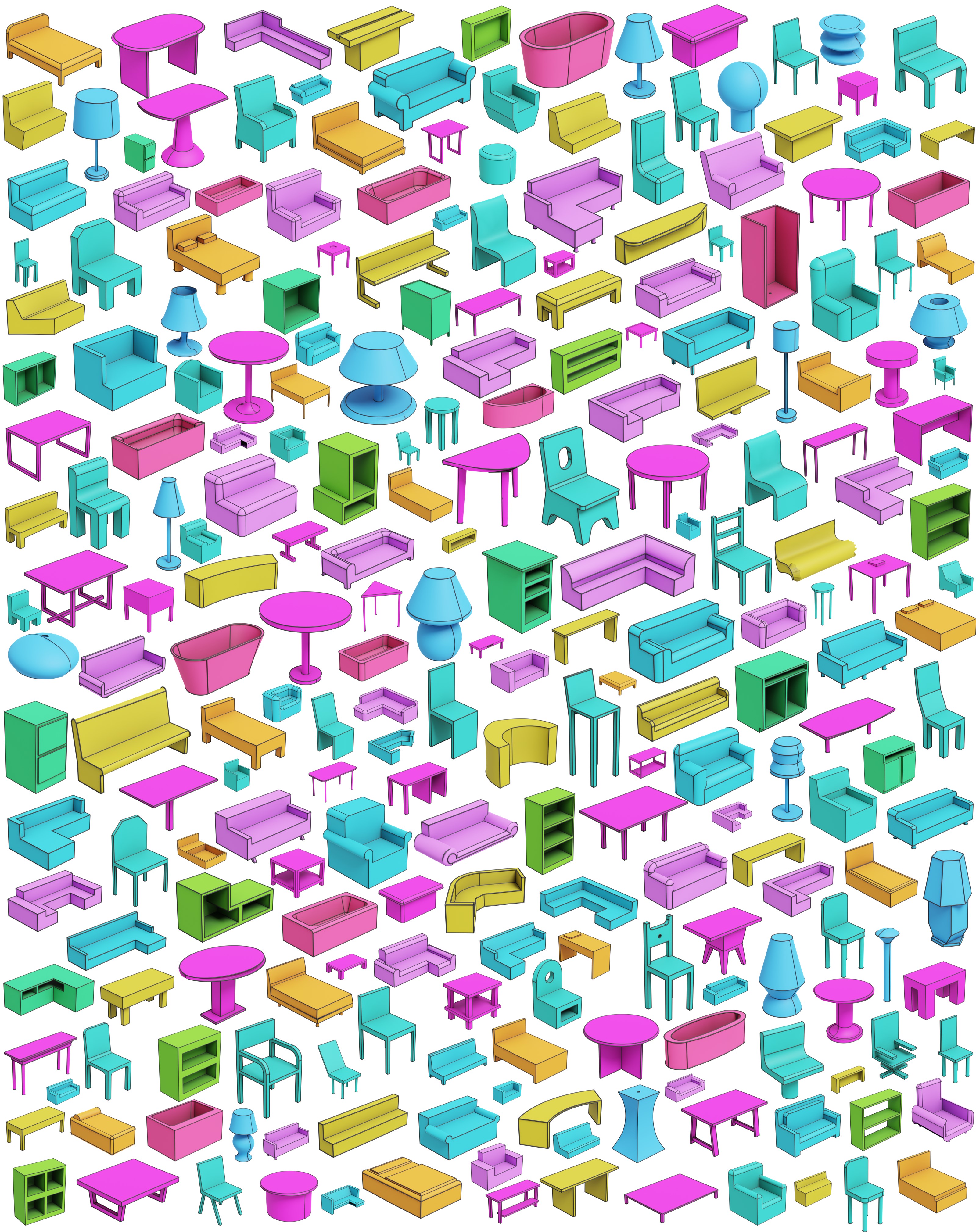}
        \caption{Conditionally generated furniture B-reps from \modelName, colored by category.}
        \label{fig:furniture_extra}
    \end{center}
\end{figure*}

\end{document}